\documentclass[10pt,twocolumn,letterpaper]{article}

\usepackage{cvpr}
\usepackage{times}
\usepackage{xspace}
\usepackage{epsfig}
\usepackage{graphicx}
\usepackage{amsmath}
\usepackage{amssymb}
\usepackage{booktabs}       
\usepackage{enumitem}
\usepackage{bm}
\usepackage{algorithm,algorithmic}
\usepackage[american]{babel}

\usepackage[pagebackref=true,breaklinks=true,letterpaper=true,colorlinks,bookmarks=false,citecolor=blue]{hyperref}

\usepackage{booktabs}
\usepackage[square,numbers,sort&compress]{natbib}
\usepackage{setspace}

\cvprfinalcopy 

\def\cvprPaperID{110} 

\usepackage[usenames, dvipsnames]{color}
\usepackage[dvipsnames]{xcolor}

\newcommand{\argmin}{\mathop{\arg\min}}

\newcommand{\x}{{\bf x}}

\newcommand{\cc}{{\bf c}}

\newcommand{\y}{{\bf y}}

\newcommand{\cev}[1]{\reflectbox{\ensuremath{\vec{\reflectbox{\ensuremath{#1}}}}}}
\usepackage{caption}

\newcommand{\feat}{\textsc{feat}\xspace}

\newcommand{\ProtoNet}{\text{ProtoNet}}

\newcommand\mypara[1]{\vspace{1mm}\noindent\textbf{#1}}

\definecolor{pink}{rgb}{0.858, 0.188, 0.478}
\definecolor{darkviolet}{rgb}{0.293, 0.0, 0.508}

\begin{document}

\title{Few-Shot Learning via Embedding Adaptation with Set-to-Set Functions}

\author{Han-Jia Ye\thanks{Work mostly done when the author was a visiting scholar at USC.} \\
	Nanjing University\\
	{\tt\small yehj@lamda.nju.edu.cn}
	\and
	Hexiang Hu\\
	USC\\
	{\tt\small hexiangh@usc.edu}
	\and 
	De-Chuan Zhan\\
	Nanjing University\\
	{\tt\small zhandc@lamda.nju.edu.cn}
	\and
	Fei Sha\thanks{On leave from USC} \\
	USC \& Google \\
	{\tt\small fsha@google.com}
}
\maketitle

\begin{abstract}
Learning with limited data is a key challenge for visual recognition. Many few-shot learning methods address this challenge by learning an instance embedding function from seen classes and apply the function to instances from unseen classes with limited labels. 
This style of transfer learning is task-agnostic: the embedding function is not learned optimally discriminative with respect to the unseen classes, where discerning among them leads to the target task.
In this paper, we propose a novel approach to adapt the instance embeddings to the target classification task with a \emph{set-to-set} function, yielding embeddings that are task-specific and are discriminative. 
We empirically investigated various instantiations of such set-to-set functions and observed the Transformer is most effective --- as it naturally satisfies key properties of our desired model. 
We denote this model as \feat (few-shot embedding adaptation w/ Transformer) and validate it on both the standard few-shot classification benchmark and four extended few-shot learning settings with essential use cases, \ie, cross-domain, transductive, generalized few-shot learning, and low-shot learning. It archived consistent improvements over baseline models as well as previous methods, and established the new state-of-the-art results on two benchmarks. 
\end{abstract}

\begin{figure*}[t!]
    \small
	\centering
    \resizebox{\linewidth}{!}{
	\begin{tabular}{cccc}
	     \fbox{\includegraphics[width=4.0cm]{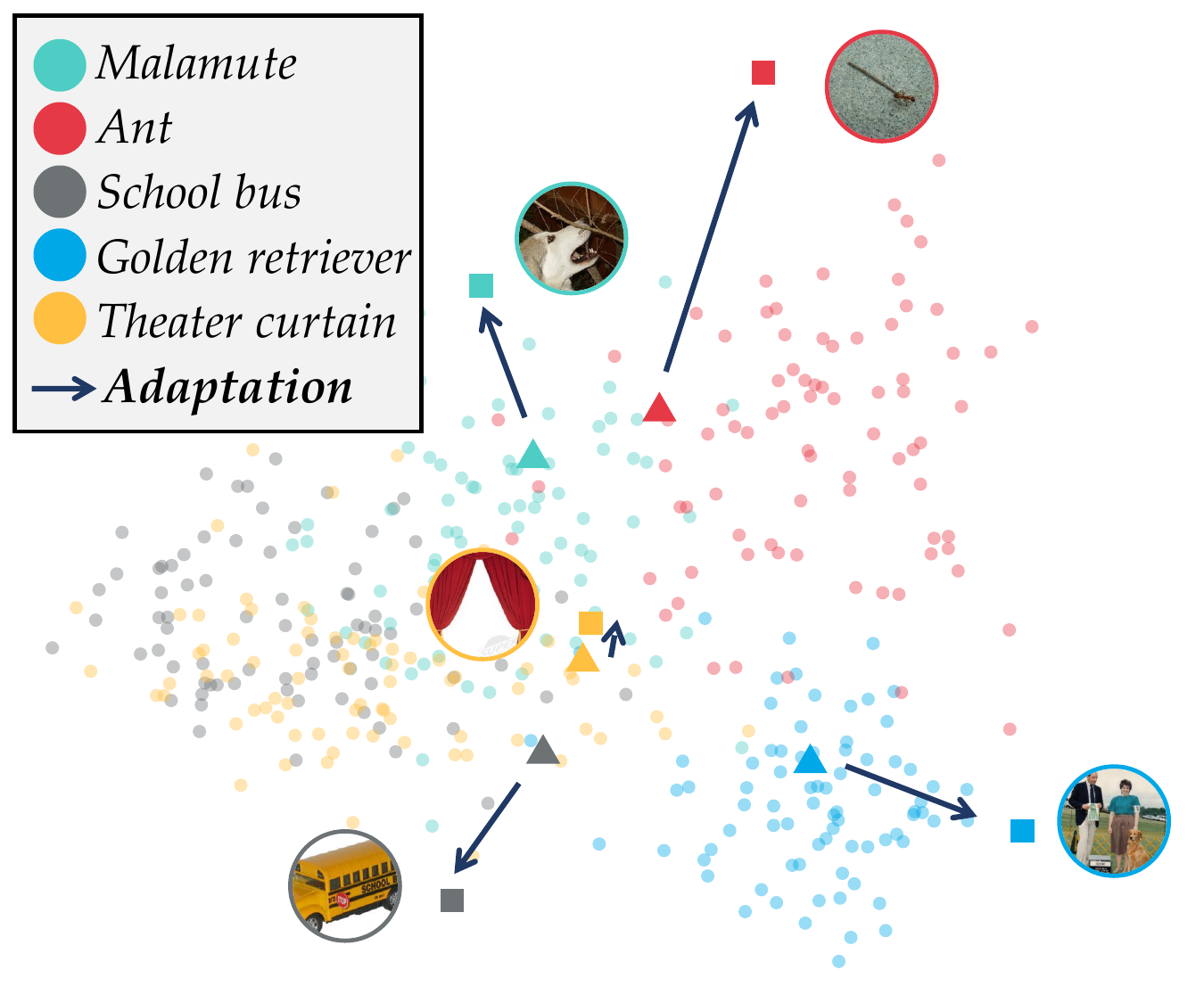}}
		& \fbox{\includegraphics[width=4.0cm]{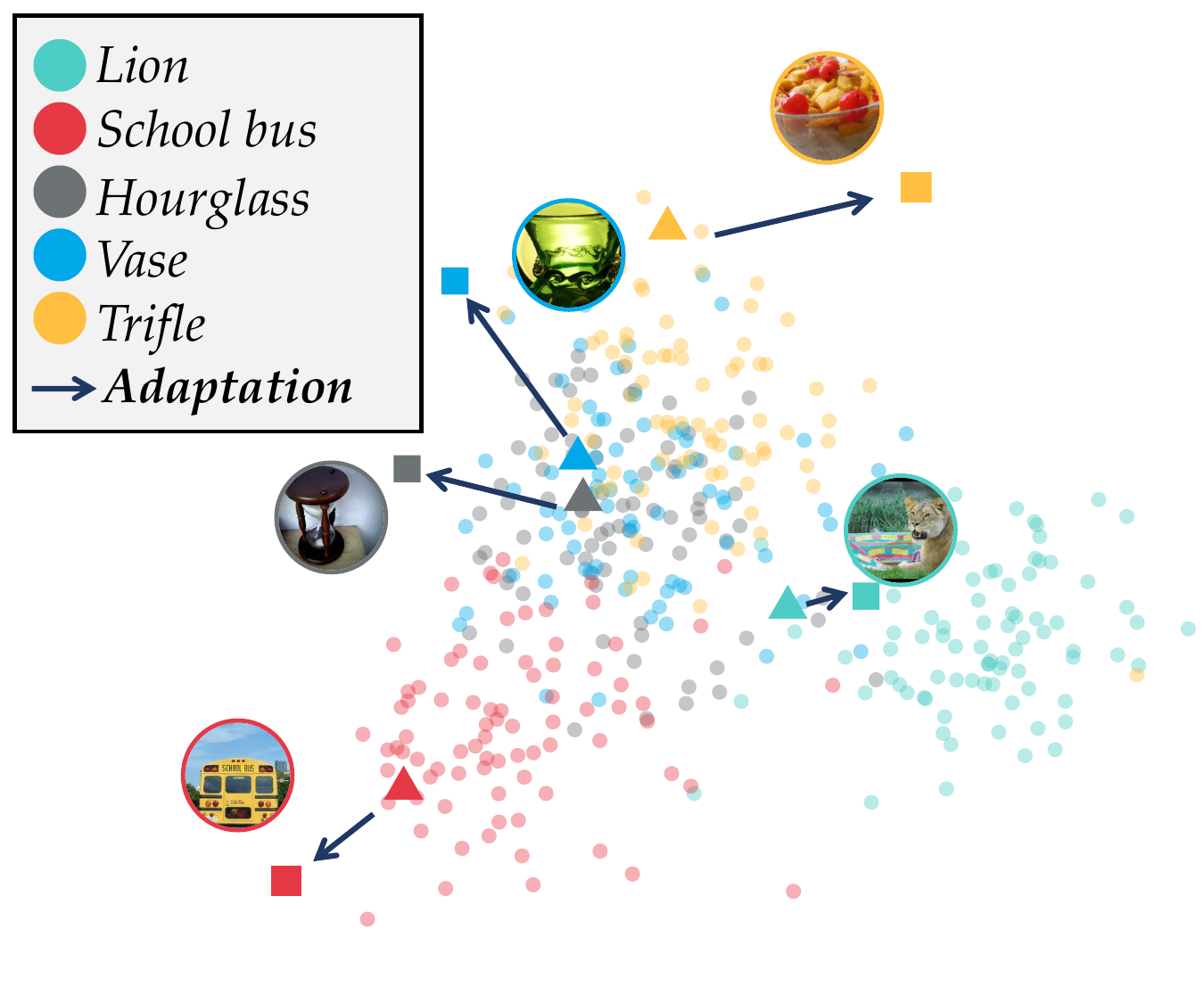}}
		& \fbox{\includegraphics[width=4.0cm]{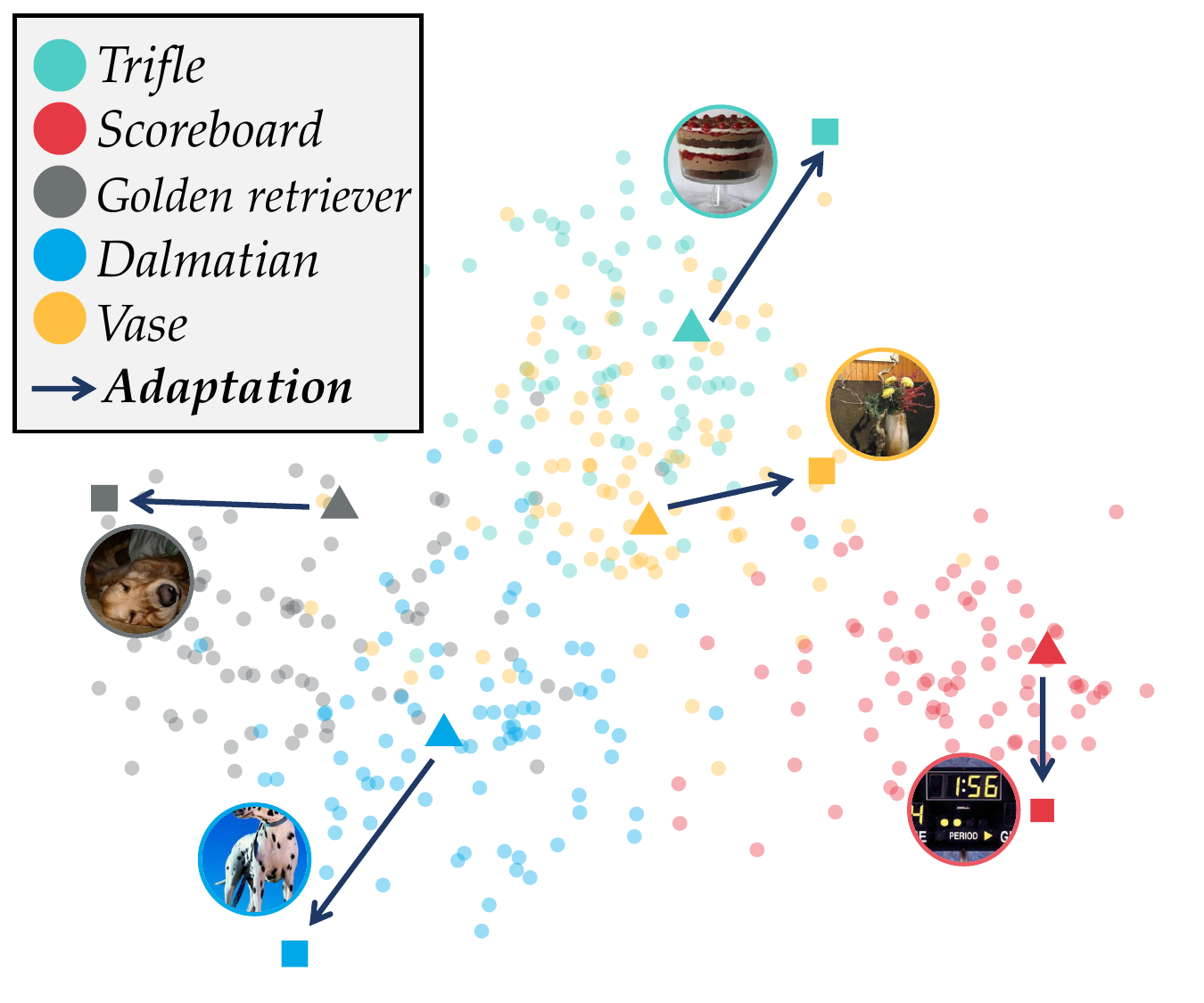}}
		& \fbox{\includegraphics[width=4.0cm]{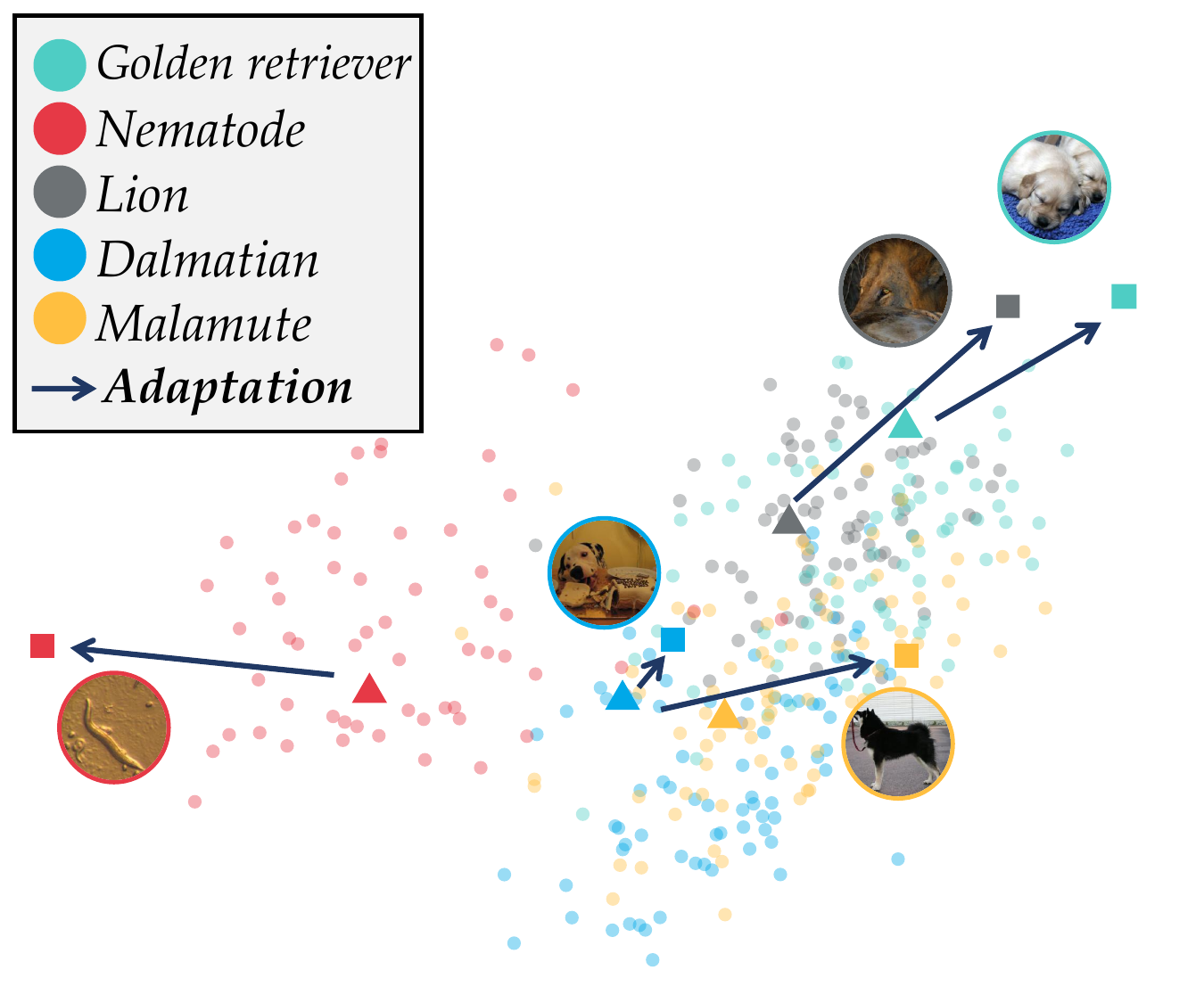}} \\
			\addlinespace
	     	\bf (a) \textcolor{red}{Acc$\xspace\uparrow$:} {$40.33\% \rightarrow 55.33\%$} 
		& \bf (b) \textcolor{red}{Acc$\xspace\uparrow$:} {$48.00\% \rightarrow 69.60\%$}
		& \bf (c) \textcolor{red}{Acc$\xspace\uparrow$:} {$43.60\% \rightarrow 63.33\%$}
		& \bf (d) \textcolor{blue}{Acc$\xspace\downarrow$:} {$56.33\% \rightarrow 47.13\%$}
	\end{tabular}}
	\caption{\textbf{Qualitative visualization} of model-based embedding adaptation procedure (implemented using \feat) on test tasks (refer to \S~\ref{sec:ablation} for more details). Each figure shows the locations of PCA projected support embeddings (class prototypes) before and after the adaptation of {\feat}. 
	Values below are the 1-shot 5-way classification accuracy before and after the the adaptation. 
	Interestingly, the embedding adaptation step of \feat pushes the support embeddings apart from the clutter and toward their own clusters, such that they can better fits the test data of its categories. (Best view in colors!)}
	\label{fig:teaser}
\end{figure*}

\section{Introduction}
\label{sec:intro}
Few-shot visual recognition~\cite{Fei-FeiFP06One,LakeSGT11One,lake2015human,VinyalsBLKW16Matching,FinnAL17Model} emerged as a promising direction in tackling the challenge of learning new visual concepts with limited annotations. Concretely, it distinguishes two sets of visual concepts: \textsc{seen} and \textsc{unseen} ones. The target task is to construct visual classifiers to identify classes from the \textsc{unseen} where each class has a very small number of exemplars (``few-shot''). The main idea is to discover transferable visual knowledge in the \textsc{seen} classes, which have ample labeled instances, and leverage it to construct the desired classifier. For example, state-of-the-art approaches for few-shot learning~\cite{VinyalsBLKW16Matching,SnellSZ17Prototypical,TriantafillouZU17Few,Rusu2018Meta} usually learn a discriminative instance embedding model on the \textsc{seen} categories, and apply it to visual data in \textsc{unseen} categories. In this common embedding space, non-parametric classifiers (\eg, nearest neighbors) are then used to avoid learning complicated recognition models from a small number of examples. 

Such approaches suffer from one important limitation. Assuming a common embedding space implies that the discovered knowledge -- discriminative visual features -- on the \textsc{seen} classes are equally effective for \emph{any} classification tasks constructed for an arbitrary set of \textsc{unseen} classes. In concrete words, suppose we have two different target tasks: discerning ``cat'' versus ``dog'' and discerning ``cat'' versus ``tiger''. Intuitively, each task uses a different set of discriminative features. Thus, the most desired embedding model first needs to be able to extract discerning features for either task at the same time. This could be a challenging aspect in its own right as the current approaches are agnostic to what those ``downstream'' target tasks are and could accidentally de-emphasize selecting features for future use. Secondly, even if both sets of discriminative features are extracted, they do not necessarily lead to the optimal performance for a \emph{specific} target task. The most useful features for discerning ``cat'' versus ``tiger'' could be irrelevant and noise to the task of discerning ``cat'' versus ``dog''!

What is missing from the current few-shot learning approaches is an \emph{adaptation} strategy that tailors the visual knowledge extracted from the \textsc{seen} classes to the \textsc{unseen} ones in a target task. In other words, we desire separate embedding spaces where each one of them is customized such that the visual features are most discriminative for a given task.
Towards this, we propose a few-shot model-based embedding adaptation method that adjusts the instance embedding models derived from the \textsc{seen} classes.
Such model-based embedding adaptation requires a \emph{set-to-set function}: a function mapping that takes all instances from the few-shot support set and outputs the set of adapted support instance embeddings, with elements in the set co-adapting with each other. Such output embeddings are then assembled as the prototypes for each visual category and serve as the nearest neighbor classifiers. Figure~\ref{fig:teaser} qualitatively illustrates the embedding adaptation procedure (as results of our best model). These class prototypes spread out in the embedding space toward the samples cluster of each category, indicating the effectiveness of embedding adaptation. 

In this paper, we implement the set-to-set transformation using a variety of function approximators, including bidirectional LSTM~\cite{HochreiterS97Long} (Bi-LSTM), deep sets~\cite{Zaheer2017Deep}, graph convolutional network (GCN)~\cite{Kipf2016Semi}, and Transformer~\cite{Lin2017SelfAttention,VaswaniNIPS17Attention}. Our experimental results (refer to \S~\ref{sec:main_result}) suggest that Transformer is the most parameter efficient choice that at the same time best implements the key properties of the desired set-to-set transformation, including {\emph{contextualization}}, {\emph{permutation invariance}}, {\emph{interpolation}} and {\emph{extrapolation}} capabilities (see \S~\ref{sec:adapt}). As a consequence, we choose the set-to-set function instantiated with Transformer to be our final model and denote it as \feat (\textbf{F}ew-shot \textbf{E}mbedding \textbf{A}daptation with \textbf{T}ransformer). We further conduct comprehensive analysis on \feat and evaluate it on many extended tasks, including few-shot domain generalization, transductive few-shot learning, and generalized few-shot learning.

\vspace{1pt}
\noindent Our overall contribution is three-fold.
\vspace{-\topsep}
\begin{itemize}[leftmargin=*]
    \setlength{\itemsep}{0pt}
    \setlength{\parskip}{3pt}
    \item We formulate the few-shot learning as a model-based embedding adaptation to make instance embeddings task-specific, via using a set-to-set transformation. 
    \item We instantiate such set-to-set transformation with various function approximators, validating and analyzing their few-shot learning ability, task interpolation ability, and extrapolation ability, \etc. It concludes our model (\feat) that uses the Transformer as the set-to-set function.
    \item We evaluate our \feat model on a variety of extended few-shot learning tasks, where it achieves superior performances compared  with strong baseline approaches. 
\end{itemize}

\section{Related Work}
\label{sec:related}
Methods specifically designed for few-shot learning fall broadly into two categories. The first is to control how a classifier for the target task should be constructed. One fruitful idea is the meta-learning framework where the classifiers are optimized \emph{in anticipation} that a future update due to data from a new task performs well on that task~\cite{AndrychowiczDCH16Learning,Sachin2017,FinnAL17Model,Antoniou2018How,Nichol2018On,Rusu2018Meta,Lee2018Gradient,Gui2018Few}, or the classifier itself is directly meta-predicted by the new task data~\cite{Qiao2017Few,Wei2018Piecewise}.

Another line of approach has focused on learning generalizable instance embeddings~\cite{changpinyo2016synthesized,akata2013label,koch2015siamese,TriantafillouZU17Few,VinyalsBLKW16Matching,Metz2018Learning,changpinyo2017predicting,Scott2018Adapted,Hsu2018Unsupervised} and uses those embeddings on simple classifiers such as nearest neighbor rules. The key assumption is that the embeddings capture all necessarily discriminative representations of data such that simple classifiers are sufficed, hence avoiding the danger of overfitting on a small number of labeled instances. Early work such as \cite{koch2015siamese} first validated the importance of embedding in one-shot learning, whilst \cite{VinyalsBLKW16Matching} proposes to learn the embedding with a soft nearest neighbor objective, following a meta-learning routine. Recent advances have leveraged different objective functions for learning such embedding models, \eg, considering the class prototypes~\cite{SnellSZ17Prototypical}, decision ranking~\cite{TriantafillouZU17Few}, and similarity comparison~\cite{Flood2017Learning}. Most recently, \cite{Garcia2017Few} utilizes the graph convolution network~\cite{Kipf2016Semi} to unify the embedding learning.

Our work follows the second school of thoughts. The main difference is that we do not assume the embeddings learned on \textsc{seen} classes, being agnostic to the target tasks, are necessarily discriminative for those tasks. In contrast, we propose to \emph{adapt} those embeddings for each target task {\em with a set-to-set function} so that the transformed embeddings are better aligned with the discrimination needed in those tasks. We show empirically that such task-specific embeddings perform better than task-agnostic ones. MetaOptNet~\cite{Lee2019Meta} and CTM~\cite{Li2019Finding} follow the same spirit of learning task-specific embedding (or classifiers) via either explicitly optimization of target task or using concentrator and projector to make distance metric task-specific.

\section{Learning Embedding for Task-agnostic FSL}
\label{sec:problem}
In the standard formulation of few-shot learning (FSL)~\cite{VinyalsBLKW16Matching,FinnAL17Model}, a task is represented as a $M$-shot $N$-way classification problem with $N$ classes sampled from a set of visual concepts $\mathcal{U}$ and $M$ (training/support) examples per class. We denote the training set (also referred as support sets in the literature) as $\mathcal{D}_{\mathbf{train}} = \{\x_{i}, \y_{i}\}_{i=1}^{NM}$, with the instance $\x_{i}\in\mathbb{R}^{D}$ and the one-hot labeling vector $\y_{i} \in \{0,1\}^N$.
We will use ``support set'' and ``training set'' interchangeably in the paper.
In FSL, $M$ is often small (\eg, $M=1$ or $M=5$). The goal is to find a function $f$ that classifies a test instance $\x_{\mathbf{test}}$ by $\hat{\y}_{\mathbf{test}} = f(\x_{\mathbf{test}};  \mathcal{D}_{\mathbf{train}})\in\{0,1\}^N$.

Given a small number of training instances, it is challenging to construct complex classifiers $f(\cdot)$. To this end, the learning algorithm is also supplied with additional data consisting of ample labeled instances. These additional data are drawn from visual classes $\mathcal{S}$, which does not overlap with $\mathcal{U}$. We refer to the original task as \emph{the target task} which discerns $N$ \textsc{unseen} classes $\mathcal{U}$. To avoid confusion, we denote the data from the \textsc{seen} classes $\mathcal{S}$ as $\mathcal{D}^{\mathcal{S}}$.

\begin{figure}[t!]
	\centering
	\small
	\includegraphics[width=0.99\linewidth]{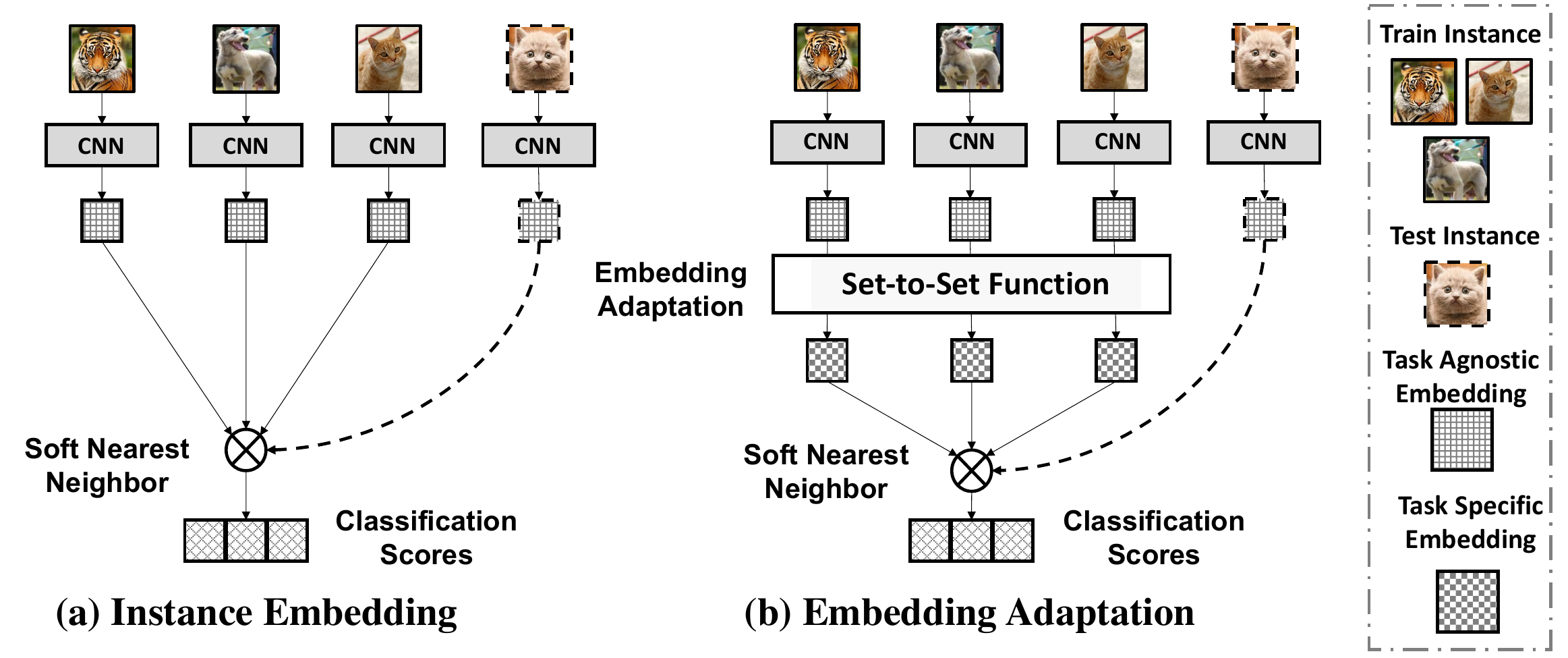}
	\caption{Illustration of the proposed \textbf{F}ew-Shot \textbf{E}mbedding \textbf{A}daptation \textbf{T}ransformer~(\feat). Existing methods usually use the same embedding function $\mathbf{E}$ for all tasks. We propose to adapt the embeddings to each target few-shot learning task with a set-to-set function such as Transformer, BiLSTM, DeepSets, and GCN.}
\label{fig:overview}
\end{figure}

To learn $f(\cdot)$ using $\mathcal{D}^{\mathcal{S}}$, we synthesize many $M$-shot $N$-way FSL tasks by sampling the data in the meta-learning manner~\cite{VinyalsBLKW16Matching,FinnAL17Model}. 
Each sampling gives rise to a task to classify a test set instance $\x^{\mathcal{S}}_{\mathbf{test}}$ into one of the $N$ \textsc{seen} classes by $f(\cdot)$, where the test instances set $\mathcal{D}^{\mathcal{S}}_{\mathbf{test}}$ is composed of the labeled instances with the same distribution as $\mathcal{D}^{\mathcal{S}}_{\mathbf{train}}$.
Formally, the function $f(\cdot)$ is learnt to minimize the averaged error over those sampled tasks
\begin{equation}
f^*= \argmin_f  \sum_{(\x^\mathcal{S}_{\mathbf{test}}, \y^\mathcal{S}_{\mathbf{test}}) \in \mathcal{D}^{\mathcal{S}}_{\mathbf{test}}}  \ell(f(\x^\mathcal{S}_{\mathbf{test}} ;  \mathcal{D}^\mathcal{S}_{\mathbf{train}}), \y^\mathcal{S}_{\mathbf{test}})\label{eq:general_obj} 
\end{equation}
where the loss $\ell(\cdot)$ measures the discrepancy between the prediction and the true label. For simplicity, we have assumed we only synthesize one task with test set $\mathcal{D}^{\mathcal{S}}_{\mathbf{test}}$. The optimal $f^*$ is then applied to the original target task.

We consider the approach based on learning embeddings for FSL~\cite{VinyalsBLKW16Matching,SnellSZ17Prototypical} (see Figure~\ref{fig:overview} (a) for an overview). In particular, the classifier $f(\cdot)$ is composed of two elements. The first is an embedding function $\phi_\x = \mathbf{E}(\x)\in\mathbb{R}^d$ that maps an instance $\x$ to a representation space. The second component applies the nearest neighbor classifiers in this space: 
\begin{align}
\hat{\y}_{\mathbf{test}} &= f(\phi_{\x_{\mathbf{test}}} ;\{\phi_{\x}, \forall (\x,\y)\in \mathcal{D}_{\mathbf{train}}\})\label{eq:softnn} \\
&\propto \exp \big( \mathbf{sim}(\phi_{\x_{\mathbf{test}}}, \phi_{\x})\big)\cdot \y , \forall (\x,\y)\in \mathcal{D}_{\mathbf{train}} \nonumber
\end{align}
Note that only the embedding function is learned by optimizing the loss in Eq.~\ref{eq:general_obj}.  For reasons to be made clear in below, we refer this embedding function as \emph{task-agnostic}.

\section{Adapting Embedding for Task-specific FSL}
\label{sec:method}
In what follows, we describe our approach for few-shot learning (FSL). We start by describing the main idea (\S~\ref{sec:adapt}, also illustrated in Figure~\ref{fig:overview}), then introduce the set-to-set adaptation function (\S~\ref{sec:transformer}). Last are learning (\S~\ref{sec:regularizer}) and implementations details (\S~\ref{sec:implementation}). 

\subsection{Adapting to Task-Specific Embeddings}
\label{sec:adapt}
\begin{algorithm}[t]
	\setstretch{1.15}
	\caption{Training strategy of embedding adaptation}
	\label{alg:FLEA}
	\begin{algorithmic}[1]{
			\REQUIRE {Seen class set $\mathcal{S}$}
			\FORALL{iteration = 1,...,MaxIteration} 
			\STATE {Sample $N$-way $M$-shot ($\mathcal{D}^{\mathcal{S}}_{\mathbf{train}}$, $\mathcal{D}^{\mathcal{S}}_{\mathbf{test}}$) from $\mathcal{S}$}
			\STATE {Compute $\phi_\x = \mathbf{E}(\x)$, for $\x \in \mathcal{X}^{\mathcal{S}}_{\mathbf{train}} \cup \mathcal{X}^{\mathcal{S}}_{\mathbf{test}}$}
			
			\FORALL{$(\x^{\mathcal{S}}_{\mathbf{test}},\y^{\mathcal{S}}_{\mathbf{test}}) \in \mathcal{D}^{\mathcal{S}}_{\mathbf{test}}$}
			    \STATE \textcolor{pink}{Compute $\{ \psi_\x\;;\forall \x\in\mathcal{X}^{\mathcal{S}}_{\mathbf{train}}\}$ with $\mathbf{T}$ via Eq.~\ref{eq:transformation}}
			    \STATE \textcolor{pink}{Predict $\hat{\y}^{\mathcal{S}}_\mathbf{test}$ with $\{ \psi_\x \}$ as Eq.~\ref{eq:feat_predict}}
			    \STATE \textcolor{pink}{Compute $\ell(\hat{\y}^{\mathcal{S}}_{\mathbf{test}}, \y^{\mathcal{S}}_{\mathbf{test}})$ with Eq.~\ref{eq:general_obj}}
			\ENDFOR
			
			\STATE {Compute $\nabla_{\mathbf{E},\mathbf{T}}\sum_{(\x^{\mathcal{S}}_{\mathbf{test}}, \y^{\mathcal{S}}_{\mathbf{test}}) \in  \mathcal{D}^{\mathcal{S}}_{\mathbf{test}}} \ell(\hat{\y}^{\mathcal{S}}_{\mathbf{test}}, \y^{\mathcal{S}}_{\mathbf{test}})$}
			\STATE {Update $\mathbf{E}$ and $\mathbf{T}$ with $\nabla_{\mathbf{E},\mathbf{T}}$ use SGD}
			\ENDFOR
		}
		\RETURN {Embedding function $\mathbf{E}$ and set function $\mathbf{T}$}.
	\end{algorithmic}
\end{algorithm}

The key difference between our approach and traditional ones is to learn \emph{task-specific} embeddings. We argue that the embedding $\phi_\x$ is not ideal. In particular, the embeddings do not necessarily highlight the most discriminative representation for a specific target task. 
To this end, we introduce an adaption step where the embedding function $\phi_\x$ (more precisely, its values on instances) is transformed. This transformation is a {\em set-to-set} function that  \textit{contextualizes} over the image instances of a set, to enable strong co-adaptation of each item. Instance functions fails to have such co-adaptation property. Furthermore, the set-to-set-function receives instances as bags, or sets without orders, requiring the function to output the set of refined instance embeddings while being \textit{permutation-invariant}. Concretely,
\begin{align}
\{\psi_\x\;;\; \forall \x \in \mathcal{X}_{\mathbf{train}}\} &= \mathbf{T}\left( \{ \phi_\x\;;\; \forall \x  \in \mathcal{X}_{\mathbf{train}} \}\right)\label{eq:transformation}\\
&=\mathbf{T}\left( \mathbf{\pi}\left\{ \phi_\x\;;\; \forall \x  \in \mathcal{X}_{\mathbf{train}} \}\right)\right)\notag
\end{align}
where $\mathcal{X}_{\mathbf{train}}$ is a set of all the instances in the training set $\mathcal{D}_{\mathbf{train}}$ for the target task. $\pi(\cdot)$ is a permutation operator over a set. Thus the set of \emph{adapted} embedding will not change if we apply a permutation over the input embedding set.
With \emph{adapted} embedding $\psi_{\x}$, the test instance $\x_{\mathbf{test}}$ can be classified by computing nearest neighbors w.r.t. $\mathcal{D}_{\mathbf{train}}$:
\begin{equation}
\hat{\y}_{\mathbf{test}} = f(\phi_{\x_{\mathbf{test}}} ; \{\psi_{\x}, \forall (\x,\y)\in \mathcal{D}_{\mathbf{train}}\})\label{eq:feat_predict}
\end{equation}

Our approach is generally applicable to different types of task-agnostic embedding function $\mathbf{E}$ and similarity measure $\mathbf{sim}(\cdot, \cdot)$, \eg, the (normalized) cosine similarity~\cite{VinyalsBLKW16Matching} or the negative distance~\cite{SnellSZ17Prototypical}. 
Both the embedding function $\mathbf{E}$ and the set transformation function $\mathbf{T}$ are optimized over synthesized FSL tasks sampled from $\mathcal{D}^\mathcal{S}$, sketched in Alg.~\ref{alg:FLEA}. Its key difference from conventional FSL is in the \textit{line 4 to line 8} where the embeddings are transformed.

\subsection{Embedding Adaptation via Set-to-set Functions}
\label{sec:transformer}
Next, we explain various choices as the instantiations of the set-to-set embedding adaptation function.

\mypara{Bidirectional LSTM (\textsc{bilstm})~\cite{HochreiterS97Long,VinyalsBLKW16Matching}} is one of the common choice to instantiate the set-to-set transformation, where the addition between the input and the hidden layer outputs of each \textsc{bilstm} cell leads to the adapted embedding.
It is notable that the output of the \textsc{bilstm} suppose to depend on the order of the input set. 
Note that using \textsc{bilstm} as embedding adaptation model is similar but different from the fully conditional embedding~\cite{VinyalsBLKW16Matching}, where the later one contextualizes both training and test instance embedding altogether, which results in a transductive setting. 

\mypara{DeepSets~\cite{Zaheer2017Deep}} is inherently a permutation-invariant transformation function. It is worth noting that \textsc{deepsets} aggregates the instances in a set into a holistic \textit{set vector}. We consider two components to implement such DeepSets transformation, an instance centric \textit{set vector} combined with a set context vector. 
For $\x \in \mathcal{X}_{\mathbf{train}}$, we define its complementary set as $\x^\complement$. Then we implement the \textsc{deepsets} by:
\begin{equation}
	\psi_{\x} = \phi_{\x} + g([\phi_{\x}; \sum_{\x_{i'}\in\x^\complement}h(\phi_{\x_{i'}})]) \label{eq:deep_set_transform}
\end{equation}
In Eq.~\ref{eq:deep_set_transform}, $g$ and $h$ are two-layer multi-layer perception (MLP) with ReLU activation which map the embedding into another space and increase the representation ability of the embedding. For each instance, embeddings in its complementary set is first combined into a {\it set vector} as the context, and then this vector is concatenated with the input embedding to obtain the residual component of adapted embedding. This conditioned embedding takes other instances in the set into consideration, and keeps the ``set (permutation invariant)'' property. In practice, we find using the maximum operator in Eq.~\ref{eq:deep_set_transform} works better than the sum operator suggested in~\cite{Zaheer2017Deep}. 

\mypara{Graph Convolutional Networks (\textsc{GCN})~\cite{Kipf2016Semi,Garcia2017Few}} propagate the relationship between instances in the set. We first construct the degree matrix $A$ to represent the similarity between instances in a set. If two instances come from the same class, then we set the corresponding element in $A$ to 1, otherwise to 0. Based on $A$, we build the ``normalized'' adjacency matrix $S$ for a given set with added self-loops $S = D^{-\frac{1}{2}}(A+I)D^{-\frac{1}{2}}$. $I$ is the identity matrix, and $D$ is the diagonal matrix whose elements are equal to the sum of elements in the corresponding row of $A+I$. 

Let $\Phi^0 = \{ \phi_\x\;;\; \forall \x  \in \mathcal{X}_{\mathbf{train}}\}$, the relationship between instances could be propagated based on $S$, \ie, 
\begin{equation}
\Phi^{t+1} = \mathbf{ReLU}(S\Phi^t W)\;,\; t=0,1,\ldots,T-1
\label{eq:gcn_transform}
\end{equation}
$W$ is a projection matrix for feature transformation. In \textsc{GCN}, the embedding in the set is transformed based on Eq.~\ref{eq:gcn_transform} multiple times, and the final $\Phi^T$ gives rise to the $\{\psi_\x\}$.

\mypara{Transformer.~\cite{VaswaniNIPS17Attention}}
We use the \textit{Transformer} architecture~\cite{VaswaniNIPS17Attention} to implement $\mathbf{T}$. In particular, we employ self-attention mechanism~\cite{VaswaniNIPS17Attention,Lin2017SelfAttention} to transform each instance embedding with consideration to its contextual instances. Note that it naturally satisfies the desired properties of $\mathbf{T}$ because it outputs refined instance embeddings and is permutation invariant. We denote it as \textbf{F}ew-Shot \textbf{E}mbedding \textbf{A}daptation with \textbf{T}ransformer~(\feat). 

Transformer is a store of triplets in the form of (query $\mathcal{Q}$, key $\mathcal{K}$, and value $\mathcal{V}$). To compute proximity and return values, those points are first linearly mapped into some space $K = W_K^\top \; \big[ \; \phi_{\x_k} ; \forall \x_k \in \mathcal{K} \; \big] \in \mathbb{R}^{d\times |\mathcal{K}|}$, which is also the same for $\mathcal{Q}$ and $\mathcal{V}$ with $W_Q$ and $W_V$ respectively. Transformer computes what is the right value for a query point --- the query $\x_q\in\mathcal{Q}$ is first matched against a list of keys $K$ where each key has a value $V$. The final value is then returned as the sum of all the values \emph{weighted} by the proximity of the key to the query point, \ie $ \psi_{\x_q} = \phi_{\x_q} + \sum_k \alpha_{qk} V_{:,k}$, where
\[
\alpha_{qk}  \propto \exp\left(\frac{\phi_{\x_q}^\top W_Q \cdot K}{\sqrt{d}}\right)
\]
and $V_{:,k}$ is the $k$-th column of $V$. In the standard FSL setup, we have $\mathcal{Q} = \mathcal{K} = \mathcal{V} =  \mathcal{X}_{\mathbf{train}}$. 

\subsection{Contrastive Learning of Set-to-Set Functions}
\label{sec:regularizer}
To facilitate the learning of embedding adaptation, we apply a contrastive objective in addition to the general one. It is designed to make sure that instances embeddings \emph{after adaptation} is similar to the same class neighbors and dissimilar to those from different classes. 
Specifically, the embedding adaptation function $\mathbf{T}$ is applied to instances of each $n$ of the $N$ class in $\mathcal{D}^{\mathcal{S}}_{\mathbf{train}}\cup\mathcal{D}^{\mathcal{S}}_{\mathbf{test}}$, which gives rise to the transformed embedding $\psi'_\x$ and class centers $\{\mathbf{c}_n\}^N_{n=1}$. Then we apply the contrastive objective to make sure training instances are close to its own class center than other centers. The total objective function (together with Eq.~\ref{eq:general_obj}) is shown as following:
\begin{align} 
\mathcal{L}(\hat{\y}_{\mathbf{test}}, \y_{\mathbf{test}})& = \ell(\hat{\y}_{\mathbf{test}}, \y_{\mathbf{test}})\\
+  \lambda \cdot \ell\big(&\mathbf{softmax}\left(\mathbf{sim}(\psi'_{\x_{\mathbf{test}}}, \mathbf{c}_n)\right), \y_{\mathbf{test}}\big)\nonumber
\end{align}
This contrastive learning makes the set transformation extract common characteristic for instances of the same category, so as to preserve the category-wise similarity. 

\subsection{Implementation details}
\label{sec:implementation}
We consider three different types of convolutional networks as the backbone for instance embedding function $\mathbf{E}$: 1) A 4-layer convolution network (ConvNet)~\cite{VinyalsBLKW16Matching,SnellSZ17Prototypical,TriantafillouZU17Few} and 2) the 12-layer residual network (ResNet) used in~\cite{Lee2019Meta}, and 3) the Wide Residual Network (WideResNet)~\cite{Zagoruyko2016WRN,Rusu2018Meta}. We apply an additional pre-training stage for the backbones over the \textsc{seen} classes, based on which our re-implemented methods are further optimized. To achieve more precise embedding, we average the same-class instances in the training set before the embedding adaptation with the set-to-set transformation. Adam~\cite{KingmaB14ADAM} and SGD are used to optimize ConvNet and ResNet variants respectively.
Moreover, we follow the most standard implementations for the four set-to-set functions --- BiLSTM~\cite{HochreiterS97Long}, DeepSets~\cite{Zaheer2017Deep}, Graph Convolutional Networks (\textsc{gcn})~\cite{Kipf2016Semi} and Transformer (\feat)~\cite{VaswaniNIPS17Attention}. We refer readers to supplementary material (SM) for complete details and ablation studies of each set-to-set functions. Our implementation is available at \url{https://github.com/Sha-Lab/FEAT}.

\section{Experiments}
\label{sec:experiment}

In this section, we first evaluate a variety of models for embedding adaptation in \S~\ref{sec:standard} with standard FSL. It concludes that \feat (with Transformer) is the most effective approach among different instantiations. Next, we perform ablation studies in \S~\ref{sec:ablation} to analyze \feat in details. Eventually, we evaluate \feat on many extended few-shot learning tasks to study its general applicability (\S~\ref{sec:extended}). This study includes {few-shot domain generalization}, {transductive few-shot learning}, {generalized few-shot learning}, and large-scale low-shot learning (refer to SM). 

\subsection{Experimental Setups}
\label{sec:ext_setup}
\mypara{Datasets.} {\it Mini}ImageNet~\cite{VinyalsBLKW16Matching} and {\it Tiered}ImageNet~\cite{Ren2018Meta} datasets are subsets of the ImageNet~\cite{RussakovskyDSKS15ImageNet}. {\it Mini}ImageNet includes a total number of 100 classes and 600 examples per class. We follow the setup provided by~\cite{Sachin2017}, and use 64 classes as \textsc{seen} categories, 16 and 20 as two sets of \textsc{unseen} categories for model validation and evaluation respectively. {\it Tiered}ImageNet is a large-scale dataset with more categories, which contains 351, 97, and 160 categories for model training, validation, and evaluation, respectively.
In addition to these, we investigate the OfficeHome~\cite{Venkateswara2017Office} dataset to validate the generalization ability of {\feat} across domains. There are four domains in OfficeHome, and two of them (``Clipart'' and ``Real World'') are selected, which contains 8722 images. After randomly splitting all classes, 25 classes serve as the \text{seen} classes to train the model, and the remaining 15 and 25 classes are used as two \textsc{unseen} for evaluation. Please refer to SM for more details. 

\mypara{Evaluation protocols.} Previous approaches~\cite{FinnAL17Model,SnellSZ17Prototypical,TriantafillouZU17Few} usually follow the original setting of~\cite{VinyalsBLKW16Matching} and evaluate the models on 600 sampled target tasks (15 test instances per class). In a later study~\cite{Rusu2018Meta}, it was suggested that such an evaluation process could potentially introduce high variances. Therefore, we follow the new and more trustworthy evaluation setting to evaluate both baseline models and our approach on 10,000 sampled tasks. We report the mean accuracy (in \%) as well as the 95\% confidence interval. 

\mypara{Baseline and embedding adaptation methods.} We re-implement the prototypical network (\ProtoNet)~\cite{SnellSZ17Prototypical} as a task-agnostic embedding baseline model. This is known as a very strong approach~\cite{Chen2019Closer} when the backbone architecture is deep, \ie, residual networks~\cite{he2016deep}. As suggested by~\cite{Oreshkin2018TADAM}, we tune the scalar temperature carefully to scale the logits of both approaches in our re-implementation. As mentioned, we implement the embedding adaptation model with four different function approximators, and denote them as \textsc{bilstm}, \textsc{deepsets}, \textsc{gcn}, and \feat (\ie Transformer). The concrete details of each model are included in the SM.

\mypara{Backbone pre-training.}
Instead of optimizing from scratch, we apply an additional pre-training strategy as suggested in~\cite{Qiao2017Few,Rusu2018Meta}. The backbone network, appended with a \textbf{softmax} layer, is trained to classify all \textsc{seen} classes with the cross-entropy loss (\eg, 64 classes in the {\it Mini}ImageNet). 
The classification performance over the penultimate layer embeddings of sampled 1-shot tasks from the model validation split is evaluated to select the best pre-trained model, whose weights are then used to initialize the embedding function $\mathbf{E}$ in the few-shot learning.

\subsection{Standard Few-Shot Image Classification}
\label{sec:standard}

\begin{table}[tbp]
    \centering
	\tabcolsep 4pt
	\small
    \caption{
    \small
    Few-shot classification accuracy on {\it Mini}ImageNet. $\bigstar$ CTM~\cite{Li2019Finding} and SimpleShot~\cite{Wang2019Simple} utilize the ResNet-18.
    (see SM for the full table with confidence intervals and WRN results.).
    }
    \begin{tabular}{@{\;}lcccc@{\;}}
    \addlinespace
    \toprule
    Setups $\rightarrow$ & \multicolumn{2}{c}{\bf 1-Shot 5-Way} & \multicolumn{2}{c}{\bf 5-Shot 5-Way} \\
    Backbone $\rightarrow$ & ConvNet & ResNet & ConvNet & ResNet\\
    \midrule
    MatchNet~\cite{VinyalsBLKW16Matching}    & 43.40 & - & 51.09 & - \\
    MAML~\cite{FinnAL17Model}                & 48.70 & - & 63.11 & - \\
    ProtoNet~\cite{SnellSZ17Prototypical}    & 49.42 & - & 68.20 & - \\
    RelationNet~\cite{Flood2017Learning}     & 51.38 & - & 67.07 & - \\
    PFA~\cite{Qiao2017Few}                   & 54.53 & 59.60 & 67.87 & 73.74 \\
    TADAM~\cite{Oreshkin2018TADAM}           & - & 58.50 & - & 76.70 \\
    MetaOptNet~\cite{Lee2019Meta}           & - & 62.64 & - & 78.63 \\
    CTM~\cite{Li2019Finding}                & - & 64.12 & - & 80.51 \\
    SimpleShot~\cite{Wang2019Simple}        & 49.69 & 62.85 & 66.92 & 80.02 \\
    \midrule
    \multicolumn{5}{@{\;}l@{\;}}{\small \bf Instance embedding} \\
    \ProtoNet                              & 52.61 & 62.39 & 71.33 & 80.53 \\
    \midrule
    \multicolumn{5}{@{\;}l@{\;}}{\small \bf Embedding adaptation} \\
    {\textsc{bilstm}} & 52.13 & 63.90 & 69.15  & 80.62 \\
    {\textsc{deepsets}} & 54.41 & 64.14 & 70.96  &  80.93 \\
    {\textsc{gcn}} & 53.25 & 64.50 & 70.59  &  81.65 \\
    \textbf{\feat} 
    & \bf 55.15 & \bf {66.78} & \bf 71.61 & \bf 82.05 \\
    \bottomrule
    \end{tabular}
    \label{tab:miniImageNet}
\end{table}

We compare our proposed {\feat} method with the instance embedding baselines as well as previous methods on the standard \textit{Mini}ImageNet~\cite{VinyalsBLKW16Matching} and \textit{Tiered}ImageNet~\cite{Ren2018Meta} benchmarks, and then perform detailed analysis on the ablated models. We include additional results with CUB~\cite{WahCUB_200_2011} dataset in SM, which shares a similar observation.

\subsubsection{Main Results}
\label{sec:main_result}

\mypara{Comparison to previous State-of-the-arts.}
Table~\ref{tab:miniImageNet} and Table~\ref{tab:tieredImageNet} show the results of our method and others on the {\it Mini}ImageNet and {\it Tiered}ImageNet. First, we observe that the best embedding adaptation method (\feat) outperforms the instance embedding baseline on both datasets, indicating the effectiveness of learning task-specific embedding space. Meanwhile, the \feat model performs significantly better than the current state-of-the-art methods on {\it Mini}ImageNet dataset. On the {\it Tiered}ImageNet, we observe that the {\ProtoNet} baseline is already better than some previous state-of-the-arts based on the 12-layer ResNet backbone. This might due to the effectiveness of the pre-training stage on the {\it Tiered}ImageNet as it is larger than {\it Mini}ImageNet and a fully converged model can be itself very effective. Based on this, all embedding adaptation approaches further improves over ProtoNet almost in all cases, with \feat achieving the best performances among all approaches. Note that here our pre-training strategy is most similar to the one used in PFA~\cite{Qiao2017Few}, while we further fine-tune the backbone. Temperature scaling of the logits influences the performance a lot when fine-tuning over the pre-trained weights. Additionally, we list some recent methods (SimpleShot~\cite{Wang2019Simple}, and CTM~\cite{Li2019Finding}) using different backbone architectures such as ResNet-18 for reference.

\begin{table}[tbp]    
	\centering
	\small
	\tabcolsep 4pt
	\caption{
		\small
		Few-shot classification accuracy and 95\% confidence interval on {\it Tiered}ImageNet with the ResNet backbone.
	}
	\begin{tabular}{lcc}
		\addlinespace
		\toprule
		Setups $\rightarrow$ & {\bf 1-Shot 5-Way} & {\bf 5-Shot 5-Way} \\
		\midrule
		ProtoNet~\cite{SnellSZ17Prototypical}  & 53.31 {\tiny $\pm$ 0.89} & 72.69 {\tiny $\pm$ 0.74} \\
		RelationNet~\cite{Flood2017Learning}  & 54.48 {\tiny $\pm$ 0.93} & 71.32 {\tiny $\pm$ 0.78} \\
		MetaOptNet~\cite{Lee2019Meta} &  65.99 {\tiny $\pm$ 0.72} &  81.56 {\tiny $\pm$ 0.63} \\
		CTM~\cite{Li2019Finding} &  68.41 {\tiny $\pm$ 0.39} &  84.28 {\tiny $\pm$ 1.73} \\
		SimpleShot~\cite{Wang2019Simple}  & 69.09 {\tiny $\pm$ 0.22} & 84.58 {\tiny $\pm$ 0.16} \\
		\midrule
		\multicolumn{3}{@{\;}l@{\;}}{\small \bf Instance embedding} \\
		\ProtoNet                   & 68.23 {\tiny $\pm$ 0.23} & 84.03 {\tiny $\pm$ 0.16} \\
		\midrule
		\multicolumn{3}{@{\;}l@{\;}}{\small \bf Embedding adaptation} \\
		\textsc{bilstm}                        &  68.14 {\tiny $\pm$ 0.23} &  84.23 {\tiny $\pm$ 0.16} \\
		\textsc{deepsets}                      &  68.59 {\tiny $\pm$ 0.24} &  84.36 {\tiny $\pm$ 0.16} \\
		\textsc{gcn}                           &  68.20 {\tiny $\pm$ 0.23} &  84.64 {\tiny $\pm$ 0.16} \\
		\textbf{\feat} 
		& \bf{70.80} {\tiny $\pm$ 0.23} & \bf{84.79} {\tiny $\pm$ 0.16} \\
		\bottomrule
	\end{tabular}
	\label{tab:tieredImageNet}
\end{table}

\mypara{Comparison among the embedding adaptation models.}
Among the four embedding adaptation methods, \textsc{bilstm} in most cases achieves the worst performances and sometimes even performs worse than ProtoNet. This is partially due to the fact that \textsc{bilstm} can not easily implement the required permutation invariant property (also shown in~\cite{Zaheer2017Deep}), which confuses the learning process of embedding adaptation. Secondly, we find that \textsc{deepsets} and \textsc{gcn} have the ability to adapt discriminative task-specific embeddings but do not achieve consistent performance improvement over the baseline {\ProtoNet} especially on {\it Mini}ImageNet with the ConvNet backbone. A potential explanation is that, such models when jointly learned with the backbone model, can make the optimization process more difficult, which leads to the varying final performances. In contrast, we observe that \feat can consistently improve {\ProtoNet} and other embedding adaptation approaches in all cases, without additional bells and whistles. 
It shows that the Transformer as a set-to-set function can implement rich interactions between instances, which provides its high expressiveness to model the embedding adaptation process. 

\begin{figure}[!t]
    \small
	\centering
	\tabcolsep 1pt
	\resizebox{1.0\linewidth}{!}{ \small
	\begin{tabular}{cccc}
	    \includegraphics[height=3.3cm]{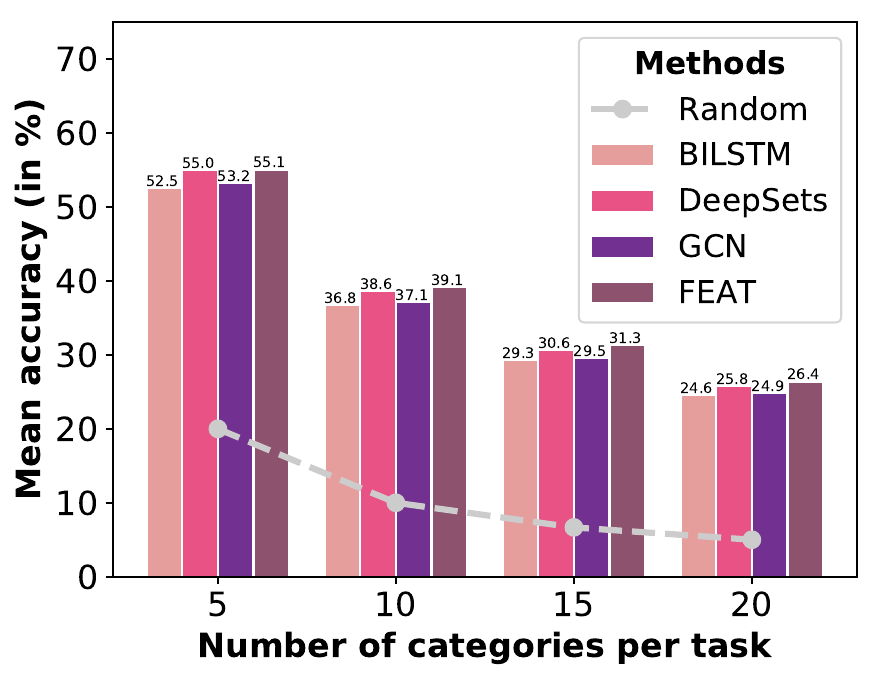}  &
	    \includegraphics[height=3.3cm]{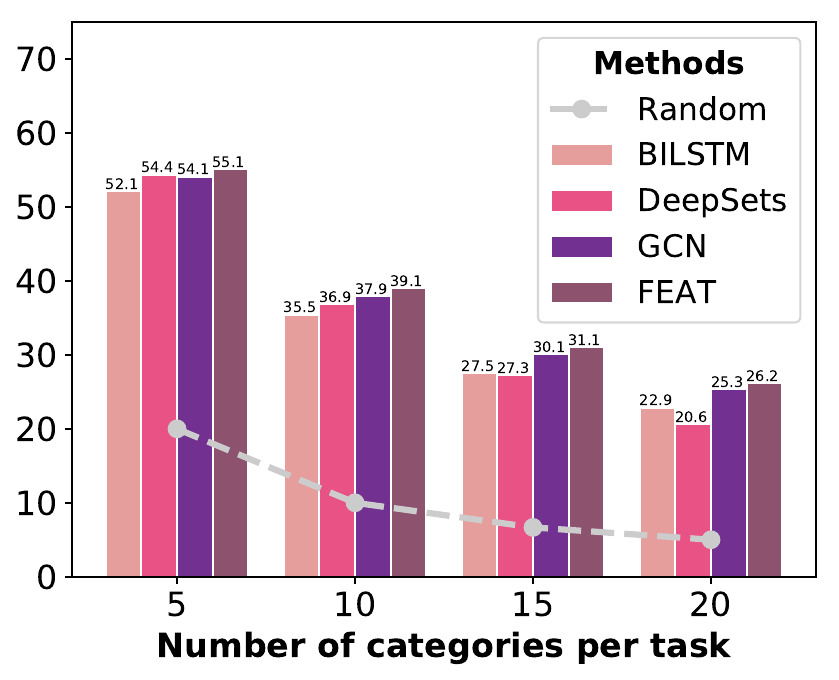}  \\
	    ({\it a}) \textbf{Way Interpolation}  & 
	    ({\it b}) \textbf{Way Extrapolation} 
	\end{tabular}
	}
	\caption{
	\small
	\textbf{Interpolation} and \textbf{Extrapolation} of few-shot tasks from the ``way'' perspective. First, We train various embedding adaptation models on 1-shot 20-way (a) or 5-way (b) classification tasks and evaluate models on unseen tasks with different number of classes (\textit{$N$=\{5, 10, 15, 20\}}). 
	It shows that \feat is superior in terms of way interpolation and extrapolation ability.
	}
	\label{fig:changeN}
\end{figure}

\begin{table}[tbp]
	\small
	\tabcolsep 5pt
	\centering
	\caption{
		\small
		Number of parameters introduced by each set-to-set function in additional to the backbone's parameters.
	}
	\begin{tabular}{@{\;}ccccc@{\;}}
		\addlinespace
		\toprule
		& \textsc{bilstm}   & \textsc{deepsets} & \textsc{gcn} & \feat \\
		\midrule
		ConvNet  & 25K  & 82K & 33K & 16K   \\
		ResNet & 2.5M  & 8.2M & 3.3M & 1.6M \\
		\bottomrule
	\end{tabular}
	\label{tab:model_size}
\end{table}

\mypara{Interpolation and extrapolation of classification ways.} Next, we study different set-to-set functions on their capability of interpolating and extrapolating across the number of classification ways. To do so, we train each variant of embedding adaptation functions with both 1-shot 20-way and 1-shot 5-way tasks, and measure the performance change as a function to the number of categories in the test time. We report the mean accuracies evaluated on few-shot classification with $N=\{5,10,15,20\}$ classes, and show results in Figure~\ref{fig:changeN}. Surprisingly, we observe that \feat \textit{achieves almost the same numerical performances in both extrapolation and interpolation scenarios}, which further displays its strong capability of learning the set-to-set transformation. Meanwhile, we observe that \textsc{deepsets} works well with interpolation but fails with extrapolation as its performance drops significantly with the larger $N$. In contrast, \textsc{gcn} achieves strong extrapolation performances but does not work as effectively in interpolation. \textsc{bilstm} performs the worst in both cases, as it is by design not permutation invariant and may have fitted an arbitrary dependency between instances.

\mypara{Parameter efficiency.} Table~\ref{tab:model_size} shows the number of additional parameters each set-to-set function has introduced. From this, we observe that with both ConvNet and ResNet backbones, \feat has the smallest number of parameters compared with all other approaches while achieving best performances from various aspects (as results discussed above), which highlights its high parameter efficiency.

All above, we conclude that: 1) learning embedding adaptation with a set-to-set model is very effective in modeling task-specific embeddings for few-shot learning 2) \feat is the most parameter-efficient function approximater that achieves the best empirical performances, together with nice permutation invariant property and strong interpolation/extrapolation capability over the classification way.

\subsubsection{Ablation Studies}
\label{sec:ablation}

We analyze {\feat} and its ablated variants on the {\it Mini}ImageNet dataset with ConvNet backbone.

\mypara{How does the embedding adaptation looks like qualitatively?} We sample four few-shot learning tasks and learn a principal component analysis (PCA) model (that projects embeddings into 2-D space) using the instance embeddings of the test data. We then apply this learned PCA projection to both the support set's pre-adapted and post-adapted embeddings. The results are shown in Figure~\ref{fig:teaser} (the beginning of the paper). In three out of four examples, post-adaptation embeddings of \feat improve over the pre-adaption embeddings. Interestingly, we found that the embedding adaptation step of \feat has the tendency of pushing the support embeddings apart from the clutter, such that they can better fit the test data of its categories. In the negative example where post-adaptation degenerates the performances, we observe that the embedding adaptation step has pushed two support embeddings ``Golden Retriever'' and ``Lion'' too close to each other. It has qualitatively shown that the adaptation is crucial to obtain superior performances and helps to contrast against task-agnostic embeddings. 

\subsection{Extended Few-Shot Learning Tasks}
\label{sec:extended}

In this section, we evaluate {\feat} on 3 different few-shot learning tasks. Specifically, cross-domain FSL, transductive FSL~\cite{Ren2018Meta,Liu2018TPN}, and generalized FSL~\cite{Chao2016Generalized}. We overview the setups briefly and please refer to SM for details.

\begin{table*}[tbp]
    \small
	\centering
	\tabcolsep 2pt
	\begin{tabular}{ccc}
	    \begin{minipage}[h]{0.28\linewidth}
        \small
        \tabcolsep 5pt
        \centering
        \begin{tabular}{@{\;}lcc@{\;}}
        	\addlinespace
        	\toprule
        	& \bf C $\rightarrow$ C & \bf C $\rightarrow$ R \\ \midrule
        	Supervised     & 34.38{\tiny $\pm$0.16} & 29.49{\tiny $\pm$0.16} \\
        	{\ProtoNet}    & 35.51{\tiny $\pm$0.16} &  29.47{\tiny $\pm$0.16} \\
        	\midrule
        	{\feat}      & \bf 36.83{\tiny $\pm$0.17} & \bf 30.89{\tiny $\pm$0.17} \\
        	\bottomrule
        \end{tabular}
	    \end{minipage} & 
    	\begin{minipage}[h]{0.36\linewidth}
    		\centering
    		\small
    		\tabcolsep 5pt
        	\begin{tabular}{@{\;}lcc@{\;}}
        		\addlinespace
        		\toprule
        		 & \bf 1-Shot & \bf 5-Shot \\
        		\midrule
        		TPN~\cite{Liu2018TPN}               & 55.51 & 69.86 \\
        		TEAM~\cite{Qiao2019Transductive}    & 56.57 & 72.04 \\
        		\midrule
        		{\feat}  & \bf 57.04 {\tiny $\pm$ 0.20} & \bf 72.89 {\tiny $\pm$ 0.16} \\
        		\bottomrule
        	\end{tabular}
        	\label{tab:transductive}
    	\end{minipage} & 
        \begin{minipage}[h]{0.36\linewidth}
    	    \small
    	    \tabcolsep 3pt
        	\begin{tabular}{@{\;}lccc@{\;}}
        		\addlinespace
        		\toprule
        		 & \bf \textsc{seen} & \bf \textsc{unseen} & \bf \textsc{combined}\\ \midrule
        		Random & 1.56 {\tiny $\pm$0.00}& 20.00{\tiny $\pm$0.00} & 1.45{\tiny $\pm$0.00}\\
        		\ProtoNet & 41.73{\tiny $\pm$0.03} & 48.64{\tiny $\pm$0.20} & 35.69{\tiny $ \pm$0.03} \\
        		\midrule
        		{{\feat}} & \bf 43.94{\tiny $\pm$0.03} & \bf 49.72{\tiny $\pm$0.20} & \bf 40.50{\tiny $\pm$0.03} \\
        		\bottomrule
        	\end{tabular}
        	\label{tab:joint}
    	\end{minipage} \\
    	\addlinespace
	    \bf (a) Few-shot domain generalization & \bf (b) Transductive few-shot learning & \bf (c) Generalized few-shot learning
	\end{tabular}
	\caption{We evaluate our model on three additional few-shot learning tasks: \textbf{(a)} Few-shot domain generalization, \textbf{(b)} Transductive few-shot learning, and \textbf{(c)} Generalized few-shot learning. We observe that \feat consistently outperform all previous methods or baselines. }
	\label{tab:cross_domain}
\end{table*}

\begin{figure}[tbp]
    \small
	\centering
		\begin{minipage}[h]{7.8cm}
		\centering \includegraphics[width=0.9\columnwidth]{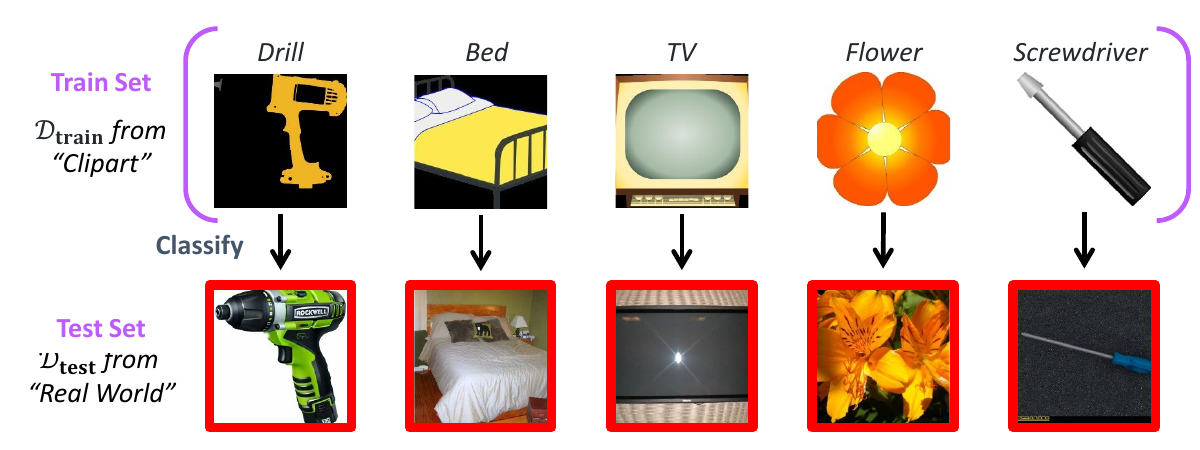}
	\end{minipage}
	\begin{minipage}[h]{7.8cm}
		\centering \includegraphics[width=0.9\columnwidth]{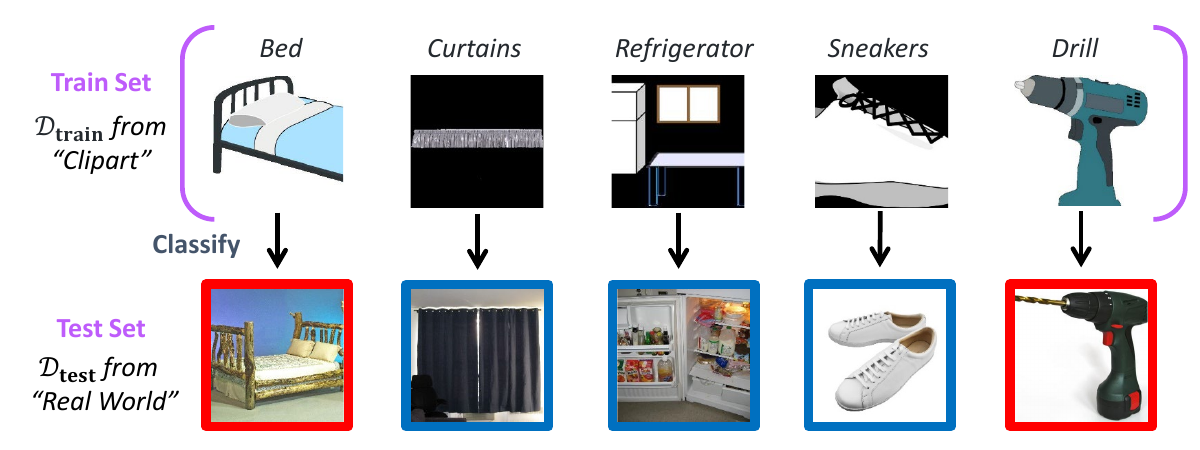}
	\end{minipage}
	\caption{Qualitative results of \textbf{few-shot domain-generalization} for {\feat}. Correctly classified examples are shown in \textbf{\textcolor{red}{red}} boxes and incorrectly ones are shown in \textbf{\textcolor{blue}{blue}} boxes. We visualize one task that \feat succeeds (top) and one that fails (bottom).}
	\label{fig:cross_domain}
\end{figure}

\mypara{FS Domain Generalization} assumes that examples in \textsc{unseen} support and test set can come from the different domains, \eg, sampled from different distributions~\cite{Dong2018Domain,Kang2018Transferable}. The example of this task can be found in Figure~\ref{fig:cross_domain}. It requires a model to recognize the intrinsic property than texture of objects, and is de facto analogical recognition.

\mypara{Transductive FSL.} The key difference between standard and transductive FSL is whether test instances arrive one at a time or all simultaneously. The latter setup allows the structure of unlabeled test instances to be utilized. Therefore, the prediction would depend on both the training (support) instances and all the available test instances in the target task from \textsc{unseen} categories.

\mypara{Generalized FSL.}
Prior works assumed the test instances coming from unseen classes only. Different from them, the \textit{generalized FSL} setting considers test instances from both \textsc{seen} and \textsc{unseen} classes~\cite{Ren2018Incremental}. In other words, during the model evaluation, while support instances all come from $\mathcal{U}$, the test instances come from $\mathcal{S}\cup\mathcal{U}$, and the classifier is required to predict on both \textsc{seen} and \textsc{unseen} categories. 
.

\vspace{-5mm}
\subsubsection{Few-Shot Domain Generalization} 
\label{sec:generalization}

We show that \feat learns to adapt \textit{the intrinsic structure of tasks}, and \textbf{generalizes across domains}, \ie, predicting test instances even when the visual appearance is changed.

\mypara{Setups.} We train the FSL model in the standard domain and evaluate with cross-domain tasks, where the $N$-categories are aligned but domains are different. In detail, a model is trained on tasks from the ``Clipart'' domain of OfficeHome dataset~\cite{Venkateswara2017Office}, then the model is required to generalize to both ``Clipart (\textbf{C})'' and``Real World (\textbf{R})'' test instances. In other words, we need to classify complex real images by seeing only a few sketches (Figure~\ref{fig:cross_domain} gives an overview of data). 

\mypara{Results.} Table~\ref{tab:cross_domain} (a) gives the quantitative results and Figure~\ref{fig:cross_domain} qualitatively examines it. Here, the ``supervised'' denotes a model trained with the standard classification strategy and then its penultimate layer's output feature is used as the nearest neighbor classifier. We observe that {\ProtoNet} can outperform this baseline on tasks when evaluating instances from ``Clipart'' but not ones from ``real world''. However, \feat improves over ``real world'' few-shot classification even only seeing the support data from ``Clipart''. 

\subsubsection{Transductive Few-Shot Learning}
\label{sec:transductive}

We show that without additional efforts in modeling, \feat outperforms existing methods in transductive FSL.  

\mypara{Setups.} We further study this semi-supervised learning setting to see how well \feat can incorporate test instances into joint embedding adaptation. Specifically, we use the unlabeled test instances to augment the key and value sets of Transformer (refer to SM for details), so that the embedding adaptation takes relationship of all test instances into consideration. We evaluate this setting on the transductive protocol of \textit{Mini}ImageNet~\cite{Ren2018Meta}. With the adapted embedding, \feat makes predictions based on Semi-ProtoNet~\cite{Ren2018Meta}. 

\mypara{Results.} We compare with two previous approaches, TPN~\cite{Liu2018TPN} and TEAM~\cite{Qiao2019Transductive}. The results are shown in Table~\ref{tab:cross_domain} (b).
We observe that \feat improves its standard FSL performance (refer to Table~\ref{tab:miniImageNet}) and also outperforms previous semi-supervised approaches by a margin. 

\subsubsection{Generalized Few-Shot Learning}
\label{sec:generalized}
We show that \feat performs well on generalized few-shot classification of both \textsc{seen} and \textsc{unseen} classes.

\mypara{Setups.} In this scenario, we evaluate not only on classifying test instances from a $N$-way $M$-shot task from \textsc{unseen} set $\mathcal{U}$, but also on all available \textsc{seen} classes from $\mathcal{S}$. To do so, we hold out 150 instances from each of the 64 seen classes in {\it Mini}ImageNet for validation and evaluation. 
Next, given a 1-shot 5-way training set $\mathcal{D}_{\mathbf{train}}$, we consider three evaluation protocols based on different class sets~\cite{Chao2016Generalized}: \textsc{unseen} measures the mean accuracy on test instances only from $\mathcal{U}$ (5-Way few-shot classification); \textsc{seen} measures the mean accuracy on test instances only from $\mathcal{S}$ (64-Way classification); \textsc{combined} measures the mean accuracy on test instances from $\mathcal{S}\cup\mathcal{U}$ (69-Way mixed classification). 

\mypara{Results.} The results can be found in Table~\ref{tab:cross_domain} (c). We observe that again {\feat} outperforms baseline {\ProtoNet}. To calibrate the prediction score on \textsc{seen} and \textsc{unseen} classes~\cite{Chao2016Generalized,Wang2018Low}, we select a constant seen/unseen class probability over the validation set, and subtract this calibration factor from seen classes' prediction score. Then we take the prediction with maximum score value after calibration.

\section{Discussion}
\label{sec:discussion}

A common embedding space fails to tailor discriminative visual knowledge for a target task especially when there are a few labeled training data.
We propose to do embedding adaptation with a set-to-set function and instantiate it with transformer (\feat), which customizes task-specific embedding spaces via a self-attention architecture.
The adapted embedding space leverages the relationship between target task training instances, which leads to discriminative instance representations.
{\feat} achieves the state-of-the-art performance on benchmarks, and its superiority can generalize to tasks like cross-domain, transductive, and generalized few-shot classifications.

\mypara{Acknowledgments.}{This work is partially supported by The National Key
R\&D Program of China (2018YFB1004300), DARPA\# FA8750-18-2-0117, NSF IIS-1065243, 1451412, 1513966/ 1632803/1833137, 1208500, CCF-1139148, a Google Research Award, an Alfred P. Sloan Research Fellowship, ARO\# W911NF-12-1-0241 and W911NF-15-1-0484, China Scholarship Council (CSC), NSFC (61773198, 61773198, 61632004), and NSFC-NRF joint research project 61861146001.}

{\small
\bibliographystyle{ieee}
\bibliography{feat}
}

\clearpage

\appendix 

\begin{center}
	\large{\bf Supplementary Material}
\end{center}

\section{Details of Baseline Methods}
\label{sec:supp-problem}
In this section, we describe two important embedding learning baselines \ie, Matching Network~(MatchNet)~\cite{VinyalsBLKW16Matching} and Prototypical Network~(ProtoNet)~\cite{SnellSZ17Prototypical}, to implement the prediction function $f(\x_{\mathbf{test}}; \mathcal{D}_{\mathbf{train}})$ in the few-shot learning framework.

\paragraph{MatchNet and ProtoNet.}
Both MatchNet and ProtoNet stress the learning of the embedding function $\mathbf{E}$ from the source task data $\mathcal{D}^\mathcal{S}$ with a meta-learning routine similar to Alg. 1 in the main text. We omit the super-script $\mathcal{S}$ since the prediction strategies can apply to tasks from both \textsc{seen} and \textsc{unseen} sets.

Given the training data $\mathcal{D}_{\mathbf{train}} = \{\x_{i}, \y_{i}\}_{i=1}^{NM}$ of an $M$-shot $N$-way classification task, we can obtain the embedding of each training instance based on the function $\mathbf{E}$:\footnote{In the following, we use $\phi(\x_i)$ and $\phi_{\x_i}$ exchangeably to represent the embedding of an instance $\x_i$ based on the mapping $\phi$.}
\begin{equation}
    \phi(\x_i) = \mathbf{E}(\x_i),\;\forall \x_i\in\mathcal{X}_{\mathbf{train}}
\end{equation}
To classify a test instance $\x_{\mathbf{test}}$, we perform the nearest neighbor classification , \ie,
\begin{align}
    &\hat{\y}_{\mathbf{test}} \propto \exp \big(\gamma\cdot\mathbf{sim}(\phi_{\x_{\mathbf{test}}}, \phi_{\x_i})\big)\cdot \y_i \\
    &=\;\frac{\exp \big(\gamma\cdot\mathbf{sim}(\phi_{\x_{\mathbf{test}}} \; , \phi_{\x_i})\big)}{\sum_{\x_{i'}\in\mathcal{X}_{\mathbf{train}}}\exp \big(\gamma\cdot\mathbf{sim}(\phi_{\x_{\mathbf{test}}} \;, \phi_{\x_{i'}})\big)} \cdot \y_i\nonumber \\
    &=\;\sum_{(\x_i,\y_i)\in\mathcal{D}_{\mathbf{train}}}\frac{\exp \big(\gamma\cdot\mathbf{sim}(\phi_{\x_{\mathbf{test}}} \; , \phi_{\x_i})\big)}{\sum_{\x_{i'}\in\mathcal{X}_{\mathbf{train}}}\exp \big(\gamma\cdot\mathbf{sim}(\phi_{\x_{\mathbf{test}}} \;, \phi_{\x_{i'}})\big)} \cdot \y_i\nonumber
\end{align}
Here, MatchNet finds the most similar training instance to the test one,  and assigns the label of the nearest neighbor to the test instance.
Note that $\mathbf{sim}$ represents the cosine similarity, and $\gamma > 0$ is the scalar temperature value over the similarity score, which is found important empirically~\cite{Oreshkin2018TADAM}. During the experiments, we tune this temperature value carefully, ranging from the reciprocal of $\{0.1, 1, 16, 32, 64, 128\}$.\footnote{In experiments, we find the temperature scale over logits influences the model training a lot when we optimize based on pre-trained weights.} 

The ProtoNet has two key differences compared with the MatchNet. First, when $M > 1$ in the target task, ProtoNet computes the mean of the same class embeddings as the class center (prototype) in advance and classifies a test instance by computing its similarity to the nearest class center (prototype). 
In addition, it uses the negative distance rather than the cosine similarity as the similarity metric: 
\begin{align}
    \cc_n &= \frac{1}{M}\sum_{\y_i = n} \phi(\x_i),\;\forall n = 1,\ldots,N\label{supp-eq:center}\\
    \hat{\y}_{\mathbf{test}} &\propto \exp \big(\gamma\cdot\|\phi_{\x_{\mathbf{test}}} -  \cc_{n}\|_2^2\big)\cdot \y_n \nonumber\\
    &=\sum_{n=1}^N\frac{\exp \big(-\gamma\|\phi_{\x_{\mathbf{test}}} - \cc_n\|_2^2\big)}{\sum_{n'=1}^N \exp \big(-\gamma\|\phi_{\x_{\mathbf{test}}} -  \cc_{n'}\|_2^2\big)}\y_n\label{supp-eq:proto_rule}
\end{align}
Similar to the aforementioned scalar temperature for MatchNet, in Eq.~\ref{supp-eq:proto_rule} we also consider the scale $\gamma$. Here we abuse the notation by using $\y_i=n$ to enumerate the instances with label $n$, and denote $\y_n$ as the one-hot coding of the $n$-th class. Thus Eq.~\ref{supp-eq:proto_rule} outputs the probability to classify $\x_{\mathbf{test}}$ to the $N$ classes.

In the experiments, we find ProtoNet incorporates better with {\feat}. When there is more than one shot in each class, we average all instances per class in advance by Eq.~\ref{supp-eq:center} before inputting them to the set-to-set transformation. This pre-average manner makes more precise embedding for each class and facilitates the ``downstream'' embedding adaptation. We will validate this in the additional experiments.

\begin{figure*}[!t]
	\centering
	\begin{minipage}[h]{1\textwidth}
		\centering
		\includegraphics[width=1\textwidth]{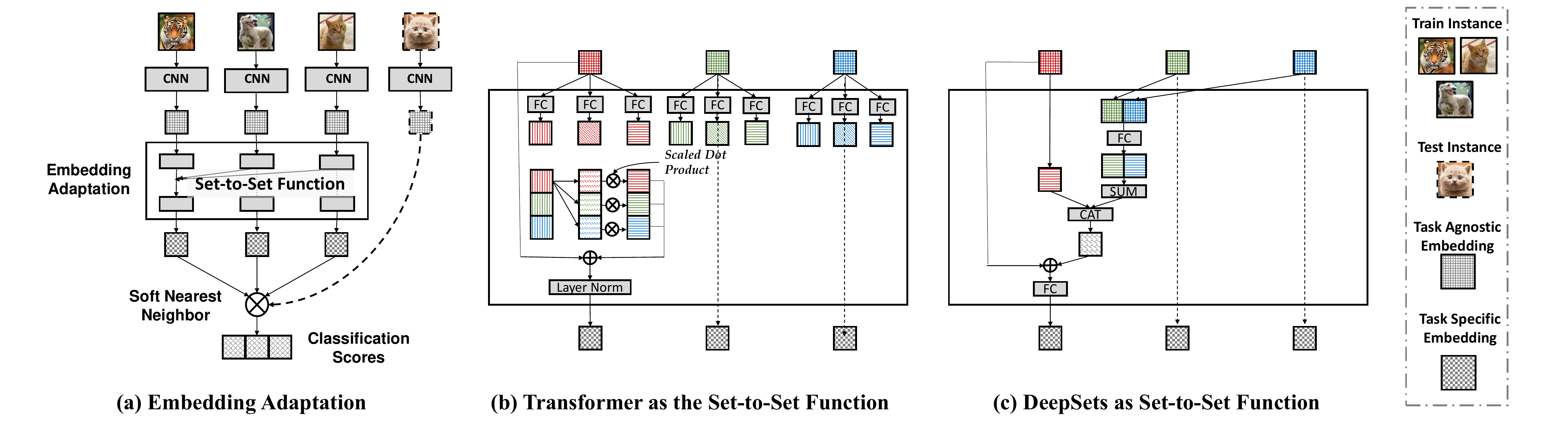}\\
	\end{minipage}
	\caption{Illustration of two embedding adpatation methods considered in the paper. (a) shows the main flow of Few-Shot Embedding Adaptation, while (b) and (c) demonstrate the workflow of Transformer and DeepSets respectively.}\label{fig:supp-demonstration}
\end{figure*}

\section{Details of the Set-to-Set Functions}
\label{sec:supp-method}
In this section, we provide details about four implementations of the set-to-set embedding adaptation function $\mathbf{T}$, \ie, the \textsc{bilstm}, \textsc{deepsets}, \textsc{gcn}, and the \textsc{transformer}. The last one is the key component in our \textbf{F}ew-shot \textbf{E}mbedding \textbf{A}daptation with \textbf{T}ransformer~(\textsc{feat}) approach. Then we will introduce the configuration of the multi-layer/multi-head transformer, and the setup of the transformer for the transductive Few-Shot Learning~(FSL).

\subsection{BiLSTM as the Set-to-Set Transformation}
Bidirectional LSTM (\textsc{bilstm})~\cite{HochreiterS97Long,VinyalsBLKW16Matching} is one of the common choice to instantiate the set-to-set transformation, where the addition between the input and the hidden layer outputs of each \textsc{bilstm} cell leads to the adapted embedding. In detail, we have
\begin{equation}
\{\cev{\phi}(\x), \vec{\phi}(\x)\}  = \textbf{BILSTM}(\{\phi(\x)\});\;\; \forall \x  \in \mathcal{X}_{\mathbf{train}}
\end{equation}
Where $\cev{\phi}(\x)$ and $\vec{\phi}(\x)$ are the hidden layer outputs of the two LSTM models for each instance embedding in the input set. Then we get the adapted embedding as 
\begin{equation}
\psi(\x) = \phi(\x) + \cev{\phi}(\x) + \vec{\phi}(\x)
\end{equation}
It is notable that the output of the \textsc{bilstm} suppose to depend on the order of the input set. 
Vinyals~\etal~\cite{VinyalsBLKW16Matching} propose to use the Fully Conditional Embedding to encode the context of both the test instance and the support set instances based on \textsc{bilstm} and LSTM w/ Attention module. Different from~\cite{VinyalsBLKW16Matching}, we apply the set-to-set embedding adaptation only over the support set, which leads to a fully inductive learning setting.

\subsection{DeepSets as the Set-to-Set Transformation}\label{sec:supp-deepsets}
Deep sets~\cite{Zaheer2017Deep} suggests a generic aggregation function over a set should be the transformed sum of all elements in this set. Therefore, a very simple set-to-set transformation baseline involves two components, an instance centric representation combined with a set context representation. 
For any instance $\x \in \mathcal{X}_{\mathbf{train}}$, we define its complementary set as $\x^\complement$. Then we implement the set transformation by:
\begin{equation}
\psi(\x) = \phi(\x) + g([\phi(\x); \sum_{\x_{i'}\in\x^\complement}h(\phi(\x_{i'}))]) \label{supp-eq:deep_set_transform}
\end{equation}
In Eq.~\ref{supp-eq:deep_set_transform}, $g$ and $h$ are transformations which map the embedding into another space and increase the representation ability of the embedding. Two-layer multi-layer perception (MLP) with ReLU activation is used to implement these two mappings. For each instance, embeddings in its complementary set are first combined into a vector as the context, and then this vector is concatenated with the input embedding to obtain the residual component of the adapted embedding. This conditioned embedding takes other instances in the set into consideration, and keeps the ``set (permutation invariant)'' property. Finally, we determine the label with the newly adapted embedding $\psi$ as Eq.~\ref{supp-eq:proto_rule}. An illustration of the DeepSets notation in the embedding adaptation can be found in Figure~\ref{fig:supp-demonstration} (c). The summation operator in Eq.~\ref{supp-eq:deep_set_transform} could also be replaced as the maximum operator, and we find the maximum operator works better than summation operator in our experiments. 

\subsection{GCN as the Set-to-Set Transformation}\label{sec:supp-gcn}
Graph Convolutional Networks (\textsc{GCN})~\cite{Kipf2016Semi,Garcia2017Few} propagate the relationship between instances in the set. We first construct a degree matrix $A\in\mathbb{R}^{NK \times NK}$ to represent the similarity between instances in a set. If two instances $\x_i$ and $\x_j$ come from the same class, then we set the corresponding element $A_{ij}$ in $A$ to 1, otherwise we have $A_{ij}=0$. 
Based on $A$, we build the ``normalized'' adjacency matrix $S$ for a given set with added self-loops $S = D^{-\frac{1}{2}}(A+I)D^{-\frac{1}{2}}$. $I\in\mathbb{R}^{NK \times NK}$ is the identity matrix, and $D$ is the diagonal matrix whose elements are equal to the sum of elements in the corresponding row of $A+I$, \ie, $D_{ii} = \sum_{j} A_{ij} + 1$ and $D_{ij}=0$ if $i\neq j$. 
Let $\Phi^0 = \{ \phi_\x\;;\; \forall \x  \in \mathcal{X}_{\mathbf{train}}\}$ be the concatenation of all the instance embeddings in the training set $\mathcal{X}_{\mathbf{train}}$. We use the super-script to denote the generation of the instance embedding matrix. The relationship between instances could be propagated based on $S$, \ie,
\begin{equation}
\Phi^{t+1} = \mathbf{ReLU}(S\Phi^t W)\;,\; t=0,1,\ldots,T-1
\label{supp-eq:gcn_transform}
\end{equation}
$W$ is a learned a projection matrix for feature transformation. In \textsc{GCN}, the embedding in the set is transformed based on Eq.~\ref{supp-eq:gcn_transform} multiple times (we propagate the embedding set two times during the experiments), and the final propagated embedding set $\Phi^T$ gives rise to the $\psi_\x$.

\subsection{Transformer as the Set-to-Set Transformation}\label{sec:supp-transformer}
In this section, we describe in details about our \textbf{F}ew-Shot \textbf{E}mbedding \textbf{A}daptation w/ \textbf{T}ransformer~(\feat) approach, specifically how to use the transformer architecture~\cite{VaswaniNIPS17Attention} to implement the set-to-set function $\mathbf{T}$, where self-attention mechanism facilitates the instance embedding adaptation with consideration of the contextual embeddings. 

As mentioned before, the transformer is a store of triplets in the form of (query, key, and value). Elements in the query set are the ones we want to do the transformation. The transformer first matches a query point with each of the keys by computing the ``query'' -- ``key'' similarities. Then the proximity of the key to the query point is used to weight the corresponding values of each key. The transformed input acts as a residual value which will be added to the input.

\paragraph{Basic Transformer.} Following the definitions in~\cite{VaswaniNIPS17Attention}, we use $\mathcal{Q}$, $\mathcal{K}$, and $\mathcal{V}$ to denote the set of the query, keys, and values, respectively. All these sets are implemented by different combinations of task instances.

To increase the flexibility of the transformer, three sets of linear projections ($W_Q\in\mathbb{R}^{d\times d'}$, $W_K\in\mathbb{R}^{d\times d'}$, and $W_V\in\mathbb{R}^{d\times d'}$) are defined, one for each set.\footnote{For notation simplicity, we omit the bias in the linear projection here.} The points in sets are first projected by the corresponding projections
\begin{equation}
\begin{array}{c c c}
Q = W_Q^\top \; \big[ \; \phi_{\x_q} ; & \forall \x_q \in \mathcal{Q} \; \big] \in \mathbb{R}^{d'\times |\mathcal{Q}|}\\[4pt]
K = W_K^\top \; \big[ \; \phi_{\x_k} ; & \forall \x_k \in \mathcal{K} \; \big] \in \mathbb{R}^{d'\times |\mathcal{K}|} \\[4pt]
V = W_V^\top \; \big[ \; \phi_{\x_v} ; & \forall \x_v \in \mathcal{V} \; \big] \in \mathbb{R}^{d'\times |\mathcal{V}|}
\end{array} 
\label{supp-eq:linear}
\end{equation}
$|\mathcal{Q}|$, $|\mathcal{K}|$, and $|\mathcal{V}|$ are the number of elements in the sets $\mathcal{Q}$, $\mathcal{K}$, and $\mathcal{V}$ respectively. Since there is a one-to-one correspondence between elements in $\mathcal{K}$ and $\mathcal{V}$ we have $|\mathcal{K}|=|\mathcal{V}|$.

The similarity between a query point $\x_q\in\mathcal{Q}$ and the list of keys $\mathcal{K}$ is then computed as ``attention'':
\begin{align}
	\alpha_{qk}  &\propto \exp\left(\frac{\phi_{\x_q}^\top W_Q \cdot K}{\sqrt{d}}\right);\; \forall \x_k \in \mathcal{K}\\
	\alpha_{q,:} &= \mathbf{softmax}\left(\frac{\phi_{\x_q}^\top W_Q \cdot K}{\sqrt{d}}\right)\in\mathbb{R}^{|\mathcal{K|}}
\end{align}The $k$-th element $\alpha_{qk}$ in the vector $\alpha_{q,:}$ reveals the particular proximity between $\x_k$ and $\x_q$. The computed attention values are then used as weights for the final embedding $\x_q$:
\begin{align}
	\tilde{\psi}_{\x_q}  & = \sum_k \alpha_{qk} V_{:,k}\\
	\psi_{\x_q}  & = \tau\big(\phi_{\x_q} + W_{\mathbf{FC}}^\top\tilde{\psi}_{\x_q}\big)
	\label{supp-eq:transformer}
\end{align}
$V_{:,k}$ is the $k$-th column of $V$. $W_{\mathbf{FC}}\in\mathbb{R}^{d'\times d}$ is the projection weights of a fully connected layer. $\tau$ completes a further transformation, which is implemented by the dropout~\cite{Srivastava2014Dropout} and layer normalization~\cite{Ba2016Layer}. The whole flow of transformer in our {\feat} approach can be found in Figure~\ref{fig:supp-demonstration} (b). With the help of transformer, the embeddings of all training set instances are adapted (we denote this approach as \feat).

\paragraph{Multi-Head Multi-Layer Transformer.}
Following~\cite{VaswaniNIPS17Attention}, an extended version of the transformer can be built with multiple parallel attention heads and stacked layers. Assume there are totally $H$ heads, the transformer {\em concatenates} multiple attention-transformed embeddings, and then uses a linear mapping to project the embedding to the original embedding space (with the original dimensionality). Besides, we can take the transformer as a feature encoder of the input query instance. Therefore, it can be applied over the input query {\em multiple times} (with different sets of parameters), which gives rise to the multi-layer transformer. We discuss the empirical performances with respect to the change number of heads and layers in \S~\ref{sec:supp-experiment}.

\subsection{Extension to transductive FSL}
\label{sec:supp-extension}

Facilitated by the flexible set-to-set transformer in Eq.~\ref{supp-eq:transformer}, our adaptation approach can naturally be extended to the transductive FSL setting. 

When classifying test instance $\x_{\mathbf{test}}$ in the transdutive scenario, other test instances $\mathcal{X}_{\mathbf{test}}$ from the $N$ categories would also be available. Therefore, we enrich the transformer's query and key/value sets
\begin{equation}
\mathcal{Q} = \mathcal{K} = \mathcal{V} = \mathcal{X}_{\mathbf{train}} \cup \mathcal{X}_{\mathbf{test}}
\end{equation}
\noindent In this manner, the embedding adaptation procedure would also consider the structure among unlabeled test instances. When the number of shots $K > 1$, we average the embedding of labeled instances in each class first before combining them with the test set embeddings.

\section{Implementation Details}
\label{sec:supp-implementation}
\paragraph{Backbone architecture.}
We consider three backbones, as suggested in the literature, as the instance embedding function $\mathbf{E}$ for the purpose of fair comparisons. We resize the input image to $84\times84\times3$ before using the backbones.

\begin{itemize}[leftmargin=*]
	\item \textbf{ConvNet.} The 4-layer convolution network~\cite{VinyalsBLKW16Matching,SnellSZ17Prototypical,TriantafillouZU17Few} contains 4 repeated blocks. In each block, there is a convolutional layer with $3\times 3$ kernel, a Batch Normalization layer~\cite{IoffeS15BN}, a \textbf{ReLU}, and a Max pooling with size $2$. We set the number of convolutional channels in each block as $64$. A bit different from the literature, we add a global max pooling layer at last to reduce the dimension of the embedding. Based on the empirical observations, this will not influence the results, but reduces the computation burden of later transformations a lot.
	\item \textbf{ResNet.} We use the 12-layer residual network in~\cite{Lee2019Meta}.\footnote{The source code of the ResNet is publicly available on \url{https://github.com/kjunelee/MetaOptNet}}
	The DropBlock~\cite{Ghiasi2018Drop} is used in this ResNet architecture to avoid over-fitting. A bit different from the ResNet-12 in~\cite{Lee2019Meta}, we apply a global average pooling after the final layer, which leads to 640 dimensional embeddings.\footnote{We use the ResNet backbone with input image size $80\times80\times3$ from~\cite{Qiao2017Few} in the old version of our paper~\cite{Ye2018Learning}, whose source code of ResNet is publicly available on \url{https://github.com/joe-siyuan-qiao/FewShot-CVPR}. Empirically we find the ResNet-12~\cite{Lee2019Meta} works better than our old ResNet architecture.}  
	\item \textbf{WRN.} We also consider the Wide residual network~\cite{Zagoruyko2016WRN,Rusu2018Meta}. We use the WRN-28-10 structure as in~\cite{Qiao2017Few,Rusu2018Meta}, which sets the depth to 28 and width to 10. After a global average pooling in the last layer of the backbone, we get a $640$ dimensional embedding for further prediction.
\end{itemize}

\paragraph{Datasets.} 
Four datasets, {\it Mini}ImageNet~\cite{VinyalsBLKW16Matching}, {\it Tiered}ImageNet~\cite{Ren2018Meta}, Caltech-UCSD Birds (CUB) 200-2011~\cite{WahCUB_200_2011}, and OfficeHome~\cite{Venkateswara2017Office} are investigated in this paper. 
Each dataset is split into three parts based on different non-overlapping sets of classes, for model training (a.k.a. meta-training in the literature), model validation (a.k.a. meta-val in the literature), and model evaluation (a.k.a. meta-test in the literature).
The CUB dataset is initially designed for fine-grained
classification. It contains in total 11,788 images of birds
over 200 species. On CUB, we randomly sampled 100
species as \textsc{seen} classes, another two 50 species are used as two \textsc{unseen} sets for model validation and evaluation~\cite{TriantafillouZU17Few}. For all images in the CUB dataset, we use the provided bounding box to crop the images as a pre-processing~\cite{TriantafillouZU17Few}. Before input into the backbone network, all images in the dataset are resized based on the requirement of the network.

\paragraph{Pre-training strategy.} As mentioned before, we apply an additional pre-training strategy as suggested in~\cite{Qiao2017Few,Rusu2018Meta}. 
The backbone network, appended with a \textbf{softmax} layer, is trained to classify all classes in the \textsc{seen} class split (\eg, 64 classes in the {\it Mini}ImageNet) with the cross-entropy loss. 
In this stage, we apply image augmentations like random crop, color jittering, and random flip to increase the generalization ability of the model.
After each epoch, we validate the performance of the pre-trained weights based on its few-shot classification performance on the model validation split. Specifically, we randomly sample 200 1-shot $N$-way few-shot learning tasks ($N$ equals the number of classes in the validation split, \eg., 16 in the {\it Mini}ImageNet), which contains 1 instance per class in the support set and 15 instances per class for evaluation. Based on the penultimate layer instance embeddings of the pre-trained weights, we utilize the nearest neighbor classifiers over the few-shot tasks and evaluate the quality of the backbone. We select the pre-trained weights with the best few-shot classification accuracy on the validation set.
The pre-trained weights are used to initialize the embedding backbone $\mathbf{E}$, and the weights of the whole model are then optimized together during the model training.

\begin{table*}[tbp]    
	\centering
	\tabcolsep 4pt
	\caption{Few-shot classification accuracy$\pm$ $95\%$ confidence interval on {\it Mini}ImageNet with ConvNet and ResNet backbones. Our implementation methods are measured over 10,000 test trials.}
	{\small
		\begin{tabular}{@{\;}lcccc @{\;}}
			\addlinespace
			\toprule
			Setups $\rightarrow$ & \multicolumn{2}{c}{\bf 1-Shot 5-Way} & \multicolumn{2}{c}{\bf 5-Shot 5-Way} \\
			Backbone Network $\rightarrow$ & ConvNet & ResNet & ConvNet & ResNet \\
			\midrule
			MatchNet~\cite{VinyalsBLKW16Matching}    & 43.40{\tiny $\pm$ 0.78} & - & 51.09{\tiny $\pm$ 0.71} & - \\
			MAML~\cite{FinnAL17Model}                & 48.70{\tiny $\pm$ 1.84} & - & 63.11{\tiny $\pm$ 0.92} & - \\
			ProtoNet~\cite{SnellSZ17Prototypical}    & 49.42{\tiny $\pm$ 0.78} & - & 68.20{\tiny $\pm$ 0.66} & - \\
			RelationNet~\cite{Flood2017Learning}     & 51.38{\tiny $\pm$ 0.82} & - & 67.07{\tiny $\pm$ 0.69} & - \\
			PFA~\cite{Qiao2017Few}                   & 54.53{\tiny $\pm$ 0.40} & - & 67.87{\tiny $\pm$ 0.20} & - \\
			TADAM~\cite{Oreshkin2018TADAM}           & - & 58.50{\tiny $\pm$ 0.30} & - & 76.70{\tiny $\pm$ 0.30} \\
			MetaOptNet~\cite{Lee2019Meta}           & - & 62.64{\tiny $\pm$ 0.61} & - & 78.63{\tiny $\pm$ 0.46} \\
			\midrule
			\multicolumn{5}{@{\;}l@{\;}}{\small \bf Baselines} \\
			MAML     & 49.24{\tiny $\pm$ 0.21} & 58.05{\tiny $\pm$ 0.10} & 67.92{\tiny $\pm$ 0.17} & 72.41{\tiny $\pm$ 0.20} \\
			MatchNet                              & 52.87{\tiny $\pm$ 0.20} & 65.64{\tiny $\pm$ 0.20} & 67.49{\tiny $\pm$ 0.17} & 78.72{\tiny $\pm$ 0.15} \\
			ProtoNet                              & 52.61{\tiny $\pm$ 0.20} & 62.39{\tiny $\pm$ 0.21} & 71.33{\tiny $\pm$ 0.16} & 80.53{\tiny $\pm$ 0.14} \\
			\midrule
			\multicolumn{5}{@{\;}l@{\;}}{\small \bf Embedding Adaptation} \\
			{\textsc{bilstm}}      & 52.13{\tiny $\pm$ 0.20} & 63.90{\tiny $\pm$ 0.21} & 69.15{\tiny $\pm$ 0.16} & 80.63{\tiny $\pm$ 0.14} \\
			\textsc{deepsets} & 54.41{\tiny $\pm$ 0.20} & 64.14{\tiny $\pm$ 0.22} & 70.96{\tiny $\pm$ 0.16} & 80.93{\tiny $\pm$ 0.14} \\ 
			{\textsc{gcn}} & 53.25{\tiny $\pm$ 0.20} & 64.50{\tiny $\pm$ 0.20} & 70.59{\tiny $\pm$ 0.16}  &  81.65{\tiny $\pm$ 0.14} \\
			\midrule
			{\bf Ours}: {\feat}      & \bf 55.15{\tiny $\pm$ 0.20} & {\bf 66.78{\tiny $\pm$ 0.20}} & {\bf 71.61{\tiny $\pm$ 0.16} } & \bf 82.05{\tiny $\pm$ 0.14} \\
			\bottomrule
	\end{tabular}}
	\label{supp-tab:miniImageNet}
\end{table*}

\paragraph{Transformer Hyper-parameters.} We follow the architecture as presented in~\cite{VaswaniNIPS17Attention} to build our \feat model. The hidden dimension $d'$ for the linear transformation in our \feat model is set to 64 for ConvNet and 640 for ResNet/WRN. The dropout rate in transformer is set as $0.5$. We empirically observed that the shallow transformer (with one set of projection and one stacked layer) gives the best overall performance (also studied in \S~\ref{sec:supp-ablation}). 

\paragraph{Optimization.} 
Following the literature, different optimizers are used for the backbones during the model training. For the ConvNet backbone, stochastic gradient descent with Adam~\cite{KingmaB14ADAM} optimizer is employed, with the initial learning rate set to be $0.002$. 
For the ResNet and WRN backbones, vanilla stochastic gradient descent with Nesterov acceleration is used with an initial rate of $0.001$. We fix the weight decay in SGD as 5e-4 and momentum as 0.9.
The schedule of the optimizers is tuned over the validation part of the dataset. 
As the backbone network is initialized with the pre-trained weights, we scale the learning rate for those parameters by $0.1$. 

\section{Additional Experimental Results}
\label{sec:supp-experiment}
\begin{table}[tbp]    
	\centering
	\tabcolsep 10pt
	\caption{Few-shot classification performance with \textbf{Wide ResNet (WRN)-28-10 backbone} on {\it Mini}ImageNet dataset (mean accuracy$\pm$95\% confidence interval). Our implementation methods are measured over 10,000 test trials.}
	{\small
		\begin{tabular}{@{\;}lcc@{\;}}
			\addlinespace
			\toprule
			Setups $\rightarrow$ & {\bf 1-Shot 5-Way} & {\bf 5-Shot 5-Way} \\    \midrule
			PFA~\cite{Qiao2017Few} & 59.60{\tiny $\pm$ 0.41} & 73.74{\tiny $\pm$ 0.19} \\
			LEO~\cite{Rusu2018Meta} & 61.76{\tiny $\pm$ 0.08} & 77.59{\tiny $\pm$ 0.12} \\
			SimpleShot~\cite{Wang2019Simple}& 63.50{\tiny $\pm$ 0.20} & 80.33{\tiny $\pm$ 0.14}\\
			ProtoNet (Ours)  & 62.60{\tiny $\pm$ 0.20} & 79.97{\tiny $\pm$ 0.14} \\
			\midrule
			{\bf Ours}: {\feat}      &  \bf 65.10 {\tiny $\pm$ 0.20} & \bf 81.11 {\tiny $\pm$ 0.14} \\
			\bottomrule
	\end{tabular}}
	\label{supp-tab:wrn}
\end{table}

\begin{table}[tbp]    
	\centering
	\tabcolsep 10pt
	\caption{Few-shot classification performance with \textbf{Wide ResNet (WRN)-28-10 backbone} on {\it Tiered}ImageNet dataset (mean accuracy$\pm$95\% confidence interval). Our implementation methods are measured over 10,000 test trials.}
	{\small
		\begin{tabular}{@{\;}lcc@{\;}}
			\addlinespace
			\toprule
			Setups $\rightarrow$ & {\bf 1-Shot 5-Way} & {\bf 5-Shot 5-Way} \\    \midrule
			LEO~\cite{Rusu2018Meta}  & 66.33{\tiny $\pm$ 0.05} & 81.44{\tiny $\pm$ 0.09} \\
			SimpleShot~\cite{Wang2019Simple}  & 69.75{\tiny $\pm$ 0.20} & \bf 85.31{\tiny $\pm$ 0.15} \\
			\midrule
			{\bf Ours}: {\feat}      &  \bf 70.41 {\tiny $\pm$ 0.23} &  84.38 {\tiny $\pm$ 0.16} \\
			\bottomrule
	\end{tabular}}
	\label{supp-tab:wrn_tiered}
\end{table}

\begin{table}[tbp]    
	\centering
	\tabcolsep 10pt
	\caption{Few-shot classification performance with ConvNet backbone on CUB dataset (mean accuracy$\pm$95\% confidence interval). Our implementation methods are measured over 10,000 test trials. }
	{\small
		\begin{tabular}{@{\;}lcc@{\;}}
			\addlinespace
			\toprule
			Setups $\rightarrow$ & {\bf 1-Shot 5-Way} & {\bf 5-Shot 5-Way} \\    \midrule
			MatchNet~\cite{VinyalsBLKW16Matching}    & 61.16 {\tiny $\pm$ 0.89} & 72.86 {\tiny $\pm$ 0.70} \\
			MAML~\cite{FinnAL17Model}                & 55.92 {\tiny $\pm$ 0.95} & 72.09 {\tiny $\pm$ 0.76} \\
			ProtoNet~\cite{SnellSZ17Prototypical}    & 51.31 {\tiny $\pm$ 0.91} & 70.77 {\tiny $\pm$ 0.69} \\
			RelationNet~\cite{Flood2017Learning}     & 62.45 {\tiny $\pm$ 0.98} & 76.11 {\tiny $\pm$ 0.69} \\ \midrule
			\multicolumn{3}{@{\;}l@{\;}}{\small \bf Instance Embedding} \\
			MatchNet                    & 67.73 {\tiny $\pm$ 0.23} & 79.00 {\tiny $\pm$ 0.16} \\
			ProtoNet                    & 63.72 {\tiny $\pm$ 0.22} & 81.50 {\tiny $\pm$ 0.15} \\
			\midrule
			\multicolumn{3}{@{\;}l@{\;}}{\small \bf Embedding Adaptation} \\
			{\textsc{bilstm}} & 62.05 {\tiny $\pm$ 0.23} & 73.51 {\tiny $\pm$ 0.19} \\
			{\textsc{deepsets}} & 67.22 {\tiny $\pm$ 0.23} & 79.65 {\tiny $\pm$ 0.16} \\
			{\textsc{GCN}} & 67.83 {\tiny $\pm$ 0.23} & 80.26 {\tiny $\pm$ 0.15} \\
			\midrule
			{\bf Ours}: {\feat}      & \bf 68.87 {\tiny $\pm$ 0.22} & \bf 82.90 {\tiny $\pm$ 0.15} \\
			\bottomrule
	\end{tabular}}
	\label{supp-tab:cub}
\end{table}
In this section, we will show more experimental results over the {\it Mini}ImageNet/CUB dataset, the ablation studies, and the extended few-shot learning.

\subsection{Main Results}
\label{sec:supp-main_result}
The full results of all methods on the {\it Mini}ImageNet can be found in Table~\ref{supp-tab:miniImageNet}. The results of MAML~\cite{FinnAL17Model} optimized over the pre-trained embedding network are also included. We re-implement the ConvNet backbone of MAML and cite the MAML results over the ResNet backbone from~\cite{Rusu2018Meta}. It is also noteworthy that the {\feat} gets the best performance among all popular methods and baselines.

We also investigate the Wide ResNet (WRN) backbone over {\it Mini}ImageNet, which is also the popular one used in~\cite{Qiao2017Few,Rusu2018Meta}. SimpleShot~\cite{Wang2019Simple} is a recent proposed embedding-based few-shot learning approach that takes full advantage of the pre-trained embeddings. We cite the results of PFA~\cite{Qiao2017Few}, LEO~\cite{Rusu2018Meta}, and SimpleShot~\cite{Wang2019Simple} from their papers.
The results can be found in Table~\ref{supp-tab:wrn}. We re-implement {ProtoNet} and our {\feat} approach with WRN. 
It is notable that in this case, our {\feat} achieves {\em much higher} promising results than the current state-of-the-art approaches. 
Table~\ref{supp-tab:wrn_tiered} shows the classification results with WRN on the {\it Tiered}ImageNet data set, where our {\feat} still keeps its superiority when dealing with 1-shot tasks.

Table~\ref{supp-tab:cub} shows the 5-way 1-shot and 5-shot classification results on the CUB dataset based on the ConvNet backbone. The results on CUB are consistent with the trend on the {\it Mini}ImageNet dataset. Embedding adaptation indeed assists the embedding encoder for the few-shot classification tasks. Facilitated by the set function property, the \textsc{deepsets} works better than the \textsc{bilstm} counterpart. Among all the results, the transformer based {\feat} gets the top tier results. 

\begin{figure}[!t]
	\small
	\centering
	\begin{minipage}[h]{4.1cm}
		\centering \includegraphics[height=3.3cm]{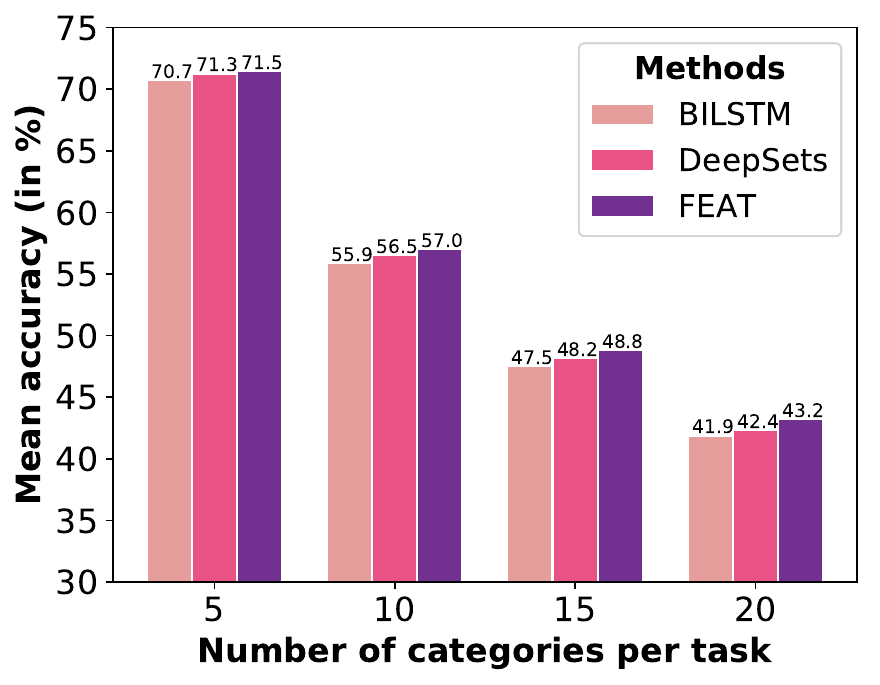}\\
		\mbox{({\it a}) \textbf{Task Interpolation}}
	\end{minipage}
	\begin{minipage}[h]{4.1cm}
		\centering \includegraphics[height=3.3cm]{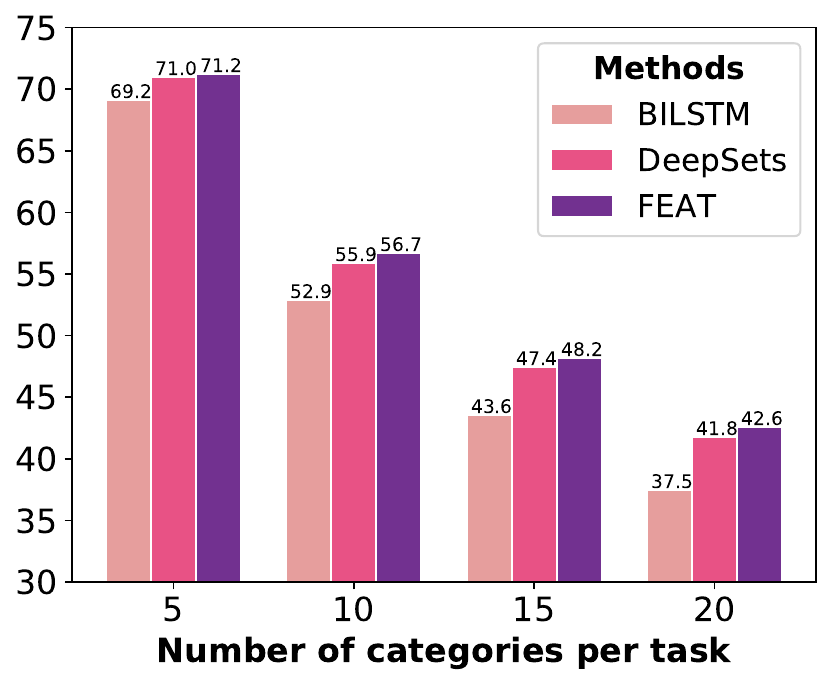}\\
		\mbox{({\it b}) \textbf{Task Extrapolation}}
	\end{minipage}
	\caption{\textbf{Interpolation} and \textbf{Extrapolation} of few-shot tasks. We train different embedding adaptation models on {\em 5-shot} 20-way or 5-way classification tasks and evaluate models on unseen tasks with different number of classes (\textit{$N$=\{5, 10, 15, 20\}}). It verifies both the interpolation and extrapolation ability of \feat on a varying number of ways in few-shot classification.}
	\label{supp-fig:changeN}
\end{figure}

\begin{table}[tbp]
	\small
	\tabcolsep 5pt
	\centering
	\caption{Ablation studies on whether the embedding adaptation improves the discerning quality of the embeddings. After embedding adaptation, \feat improves w.r.t. the before-adaptation embeddings a lot for Few-shot classification.}
	\begin{tabular}{@{\;}ccc@{\;}}
		\addlinespace
		\toprule
		& 1-Shot 5-Way   & 5-Shot 5-Way \\
		\midrule
		Pre-Adapt    & 51.60{\tiny $\pm$ 0.20}  & 70.40{\tiny $\pm$ 0.16} \\
		Post-Adapt  & 55.15{\tiny $\pm$ 0.20}  & 71.61{\tiny $\pm$ 0.16} \\
		\bottomrule
	\end{tabular}
	\label{supp-tab:adapt_ablation}
\end{table}

\begin{table}[tbp]
	\small
	\tabcolsep 5pt
	\centering
	\caption{Ablation studies on the position to average the same-class embeddings when there are multiple shots per class in {\feat} (tested on the 5-Way tasks with different numbers of shots). ``Pre-Avg'' and ``Post-Avg'' means we get the embedding center for each class before or after the set-to-set transformation, respectively.}
	\begin{tabular}{@{\;}ccc@{\;}}
		\addlinespace
		\toprule
		Setups $\rightarrow$ & Pre-Avg & Post-Avg \\
		\midrule
		5     & 71.61{\tiny $\pm$ 0.16} & 70.70{\tiny $\pm$ 0.16}  \\
		15     & 77.76{\tiny $\pm$ 0.14} & 76.58{\tiny $\pm$ 0.14} \\
		30     & 79.66{\tiny $\pm$ 0.13} & 78.77{\tiny $\pm$ 0.13} \\
		\bottomrule
	\end{tabular}
	\label{supp-tab:shot_change}
\end{table}

\begin{table}[tbp]
	\small
	\tabcolsep 5pt
	\centering
	\caption{Ablation studies on the number of heads in the Transformer of {\feat} (with number of layers fixes to one).}
	\begin{tabular}{@{\;}ccc@{\;}}
		\addlinespace
		\toprule
		Setups $\rightarrow$ & 1-Shot 5-Way & 5-Shot 5-Way \\
		\midrule
		1     & 55.15{\tiny $\pm$ 0.20} & 71.57{\tiny $\pm$ 0.16}  \\
		2     & 54.91{\tiny $\pm$ 0.20} & 71.44{\tiny $\pm$ 0.16} \\
		4     & 55.05{\tiny $\pm$ 0.20} & 71.63{\tiny $\pm$ 0.16} \\
		8     & 55.22{\tiny $\pm$ 0.20} & 71.39{\tiny $\pm$ 0.16} \\
		\bottomrule
	\end{tabular}
	\label{supp-tab:component_head}
\end{table}

\begin{table}[tbp]
	\small
	\tabcolsep 5pt
	\centering
	\caption{Ablation studies on the number of layers in the Transformer of {\feat} (with number of heads fixes to one).}
	\begin{tabular}{@{\;}ccc@{\;}}
		\addlinespace
		\toprule
		Setups $\rightarrow$ & 1-Shot 5-Way & 5-Shot 5-Way \\
		\midrule
		1     & 55.15{\tiny $\pm$ 0.20} & 71.57{\tiny $\pm$ 0.16}  \\
		2     & 55.42{\tiny $\pm$ 0.20} & 71.44{\tiny $\pm$ 0.16} \\
		3     & 54.96{\tiny $\pm$ 0.20} & 71.63{\tiny $\pm$ 0.16} \\
		\bottomrule
	\end{tabular}
	\label{supp-tab:component_layer}
\end{table}

\subsection{Ablation Studies}
\label{sec:supp-ablation}
In this section, we perform further analyses for our proposed {\feat} and its ablated variants classifying in the {ProtoNet} manner, on the {\it Mini}ImageNet dataset, using the ConvNet as the backbone network. 

\paragraph{Do the adapted embeddings improve the pre-adapted embeddings?} We report few-shot classification results by using the pre-adapted embeddings of support data (\ie, the embedding before adaptation), against those using adapted embeddings, for constructing classifiers. Table~\ref{supp-tab:adapt_ablation} shows that task-specific embeddings after adaptation improves over task-agnostic embeddings in few-shot classifications.

\paragraph{Can \feat possesses the characteristic of the set function?} We test three set-to-set transformation implementations, namely the \textsc{bilstm}, the \textsc{deepsets}, and the Transformer (\feat), w.r.t. two important properties of the set function, \ie, task interpolation and task extrapolation. In particular, the few-shot learning model is first trained with 5-shot 20-way tasks. Then the learned model is required to evaluate different 5-shot tasks with $N=\{5,10,15,20\}$ (Extrapolation). Similarly, for interpolation, the model is trained with 5-shot 20-way tasks in advance and then evaluated on the previous multi-way tasks. The classification change results can be found in Figure~\ref{supp-fig:changeN} (a) and (b). \textsc{bilstm} cannot deal with the size change of the set, especially in the task extrapolation. In both cases, {\feat} still gets improvements in all configurations of $N$.

\paragraph{When to average the same-class embeddings?}When there is more than one instance per class, \ie $M > 1$, we average the instances in the same class and use the class center to make predictions as in Eq.~\ref{supp-eq:center}. There are two positions to construct the prototypes in {\feat} --- before the set-to-set transformation (Pre-Avg) and after the set-to-set transformation (Post-Avg). In Pre-Avg, we adapt the embeddings of the centers, and a test instance is predicted based on its distance to the nearest adapted center; while in Post-Avg, the instance embeddings are adapted by the set-to-set function first, and the class centers are computed based on the adapted instance embeddings. We investigate the two choices in Table~\ref{supp-tab:shot_change}, where we fix the number of ways to 5 ($N=5$) and change the number of shots ($M$) among $\{5,15,30\}$. The results demonstrate the Pre-Avg version performs better than the Post-Avg in all cases, which shows a more precise input of the set-to-set function by averaging the instances in the same class leads to better results. So we use the Pre-Avg strategy as a default option in our experiments.

\paragraph{Will deeper and multi-head transformer help?} In our current implementation of the set-to-set transformation function, we make use of a shallow and simple transformer, \ie, one layer and one head (set of projection). From~\cite{VaswaniNIPS17Attention}, the transformer can be equipped with complex components using multiple heads and deeper stacked layers. We evaluate this augmented structure, with the number of attention heads increases to 2, 4, 8, as well as with the number of layers increases to 2 and 3. As in Table~\ref{supp-tab:component_head} and Table~\ref{supp-tab:component_layer}, we empirically observe that more complicated structures do not result in improved performance. We find that with more layers of transformer stacked, the difficulty of optimization increases and it becomes harder to train models until their convergence. Whilst for models with more heads, the models seem to over-fit heavily on the training data, even with the usage of auxiliary loss term (like the contrastive loss in our approach). It might require some careful regularizations to prevent over-fitting, which we leave for future work.

\begin{table}[tbp]
	\small
	\tabcolsep 5pt
	\centering
	\caption{Ablation studies on effects of the contrastive learning of the set-to-set function on {\feat}.}
	\begin{tabular}{@{\;}lcc@{\;}}
		\toprule
		Setups $\rightarrow$ & \bf 1-Shot 5-Way & \bf 5-Shot 5-Way \\\midrule
		$\lambda = 10$  & 53.92 {\tiny $\pm$ 0.20} & 70.41 {\tiny $\pm$ 0.16} \\
		$\lambda = 1$  & 54.84 {\tiny $\pm$ 0.20} & 71.00 {\tiny $\pm$ 0.16} \\
		$\lambda = 0.1$ & \bf 55.15 {\tiny $\pm$ 0.20} & \bf 71.61 {\tiny $\pm$ 0.16} \\
		$\lambda = 0.01$ & 54.67 {\tiny $\pm$ 0.20} & 71.26 {\tiny $\pm$ 0.16} \\
		\bottomrule
	\end{tabular}
	\label{supp-tab:reg1}
\end{table}

\paragraph{The effectiveness of contrastive loss.} 
Table~\ref{supp-tab:reg1} show the few-shot classification results with different weight values ($\lambda$) of the contrastive loss term for {\feat}. From the results, we can find that the balance of the contrastive term in the learning objective can influence the final results. Empirically, we set $\lambda=0.1$ in our experiments. 

\begin{table}[tbp]
	\small
	\tabcolsep 5pt
	\centering
	\caption{Ablation studies on the prediction strategy (with cosine similarity or euclidean distance) of {\feat}.}
	\begin{tabular}{@{\;}lcccc@{\;}}
		\addlinespace
		\toprule
		Setups $\rightarrow$ & \multicolumn{2}{c}{\bf 1-Shot 5-Way} & \multicolumn{2}{c}{\bf 5-Shot 5-Way} \\
		\midrule
		Backbone $\rightarrow$ & ConvNet & ResNet & ConvNet  & ResNet \\
		\midrule
		\multicolumn{3}{@{\;}l@{\;}}{\small \bf Cosine Similarity-based Prediction} \\
		{\feat}     & 54.64{\tiny $\pm$ 0.20} & 66.26{\tiny $\pm$ 0.20} & 71.72{\tiny $\pm$ 0.16} & 81.83{\tiny $\pm$ 0.15} \\
		\midrule
		\multicolumn{3}{@{\;}l@{\;}}{\small \bf Euclidean Distance-based Prediction} \\
		{\feat}     & 55.15{\tiny $\pm$ 0.20} & 66.78{\tiny $\pm$ 0.20} & 71.61{\tiny $\pm$ 0.16} & 82.05{\tiny $\pm$ 0.14} \\
		\bottomrule
	\end{tabular}
	\label{supp-tab:prediction_compare}
\end{table}

\paragraph{The influence of the prediction strategy.} We investigate two embedding-based prediction ways for the few-shot classification, \ie, based on the cosine similarity and the negative euclidean distance to measure the relationship between objects, respectively. We compare these two choices in Table~\ref{supp-tab:prediction_compare}. 
Two strategies in Table~\ref{supp-tab:prediction_compare} only differ in their similarity measures. In other words, with more than one shot per class in the task training set, we average the same class embeddings first, and then make classification by computing the cosine similarity or the negative euclidean distance between a test instance and a class prototype. 
During the optimization, we tune the logits scale temperature for both these methods. 
We find that using the euclidean distance usually requires small temperatures (\eg, $\gamma = \frac{1}{64}$) while a large temperature (\eg, $\gamma=1$) works well with the normalized cosine similarity. The former choice achieves a slightly better performance than the latter one.

\begin{table}[tbp]
	\small
	\tabcolsep 5pt
	\centering
	\caption{Cross-Domain 1-shot 5-way classification results of the {\feat} approach.}
	\begin{tabular}{@{\;}lccc@{\;}}
		\addlinespace
		\toprule
		& \bf C $\rightarrow$ C & \bf C $\rightarrow$ R & \bf R $\rightarrow$ R\\ 
		\midrule
		Supervised     & 34.38{\tiny $\pm$0.16} & 29.49{\tiny $\pm$0.16} & 37.43{\tiny $\pm$0.16}\\
		{ProtoNet}    & 35.51{\tiny $\pm$0.16} &  29.47{\tiny $\pm$0.16} & 37.24{\tiny $\pm$0.16}\\
		\midrule
		{\feat}      & \bf 36.83{\tiny $\pm$0.17} & \bf 30.89{\tiny $\pm$0.17} & \bf 38.49{\tiny $\pm$0.16}\\
		\bottomrule
	\end{tabular}
	\label{supp-tab:generalization1}
\end{table}

\subsection{Few-Shot Domain Generalization} 
\label{sec:supp-generalization}
We show that \feat learns to adapt \textit{the intrinsic structure of tasks}, and \textbf{generalize across domains}, \ie, predicting test instances even when the visual appearance is changed.

\par\noindent\textbf{Setups.} We train a few-shot learning model in the standard domain and evaluate it with cross-domain tasks, where the $N$-categories are aligned but domains are different. In detail, a model is trained on tasks from the ``Clipart'' domain of OfficeHome dataset~\cite{Venkateswara2017Office}, then the model is required to generalize to both ``Clipart (\textbf{C})'' and ``Real World (\textbf{R})'' instances. In other words, we need to classify complex real images by seeing only a few sketches, or even based on the instances in the ``Real World (\textbf{R})'' domain. 

\par\noindent\textbf{Results.} Table~\ref{supp-tab:generalization1} gives the quantitative results. Here, the ``supervised'' refers to a model trained with standard classification and then is used for the nearest neighbor classifier with its penultimate layer's output feature. We observe that {ProtoNet} can outperform this baseline on tasks when evaluating instances from ``Clipart'' but not ones from ``real world''. However, \feat can improve over ``real world'' few-shot classification even only seeing the support data from ``Clipart''. Besides, when the support set and the test set of the target task are sampled from the same but new domains, \eg, the training and test instances both come from ``real world'', {\feat} also improves the classification accuracy w.r.t. the baseline methods. It verifies the domain generalization ability of the {\feat} approach. 

\begin{table}[tbp]
	\centering
	\small
	\tabcolsep 5pt
	\caption{Results of models for transductive FSL with ConvNet backbone on {\it Mini}ImageNet. We cite the results of Semi-ProtoNet and TPN from~\cite{Ren2018Meta} and~\cite{Qiao2019Transductive} respectively. For TEAM~\cite{Qiao2019Transductive}, the authors do not report the confidence intervals, so we set them to 0.00 in the table. $\feat^\dagger$ and $\feat^\ddagger$ adapt embeddings with the joint set of labeled training and unlabeled test instances, while make prediction via ProtoNet and Semi-ProtoNet respectively. }
	\begin{tabular}{@{\;}lcc@{\;}}
		\addlinespace \toprule
		Setups $\rightarrow$ & \bf 1-Shot 5-Way & \bf 5-Shot 5-Way \\
		\midrule
		\multicolumn{3}{@{\;}l@{\;}}{\bf Standard} \\
		ProtoNet                            & 52.61 {\tiny $\pm$ 0.20} & 71.33 {\tiny $\pm$ 0.16} \\
		{\feat}     & 55.15 {\tiny $\pm$ 0.20} & 71.61 {\tiny $\pm$ 0.16} \\ 
		\midrule
		\multicolumn{3}{@{\;}l@{\;}}{\bf Transductive} \\
		Semi-ProtoNet~\cite{Ren2018Meta}    & 50.41 {\tiny $\pm$ 0.31} & 64.39 {\tiny $\pm$ 0.24} \\
		TPN~\cite{Liu2018TPN}               & 55.51 {\tiny $\pm$ 0.84} & 69.86 {\tiny $\pm$ 0.67} \\
		TEAM~\cite{Qiao2019Transductive}    & 56.57 {\tiny $\pm$ 0.00} & 72.04 {\tiny $\pm$ 0.00} \\
		Semi-ProtoNet (Ours)    & 55.50 {\tiny $\pm$ 0.10} & 71.76 {\tiny $\pm$ 0.08} \\
		{\feat}$^\dagger$  &  56.49 {\tiny $\pm$ 0.16} &  72.65 {\tiny $\pm$ 0.20} \\
		{\feat}$^\ddagger$  & \bf 57.04 {\tiny $\pm$ 0.16} & \bf 72.89 {\tiny $\pm$ 0.20} \\
		\bottomrule
	\end{tabular}
	\label{supp-tab:transductive}
\end{table}

\subsection{Additional Discussions on Transductive FSL}
\label{sec:supp-extended}
We list the results of the transductive few-shot classification in Table~\ref{supp-tab:transductive}, where the unlabeled test instances arrive simultaneously, so that the common structure among the unlabeled test instances could be captured.
We compare with three approaches, Semi-ProtoNet~\cite{Ren2018Meta}, TPN~\cite{Liu2018TPN}, and TEAM~\cite{Qiao2019Transductive}. Semi-ProtoNet utilizes the unlabeled instances to facilitate the computation of the class center and makes predictions similar to the prototypical network; TPN meta learns a label propagation way to take the unlabeled instances relationship into consideration; TEAM explores the pairwise constraints in each task, and formulates the embedding adaptation into a semi-definite programming form. We cite the results of Semi-ProtoNet from \cite{Ren2018Meta}, and cite the results of TPN and TEAM from \cite{Qiao2019Transductive}. We also re-implement  Semi-ProtoNet with our pre-trained backbone (the same pre-trained ConvNet weights as the standard few-shot learning setting) for a fair comparison.

In this setting, our model leverages the unlabeled test instances to augment the transformer as discussed in \S~\ref{sec:supp-transformer} and the embedding adaptation takes the relationship of all test instances into consideration. Based on the adapted embedding by the joint set of labeled training instances and unlabeled test instances, we can make predictions with two strategies. First, we still compute the center of the labeled instances, while such adapted embeddings are influenced by the unlabeled instances (we denote this approach as {\feat}$^\dagger$, which works the same way as standard {\feat} except the augmented input of the embedding transformation function); Second, we consider to take advantage of the unlabeled instances and use their adapted embeddings to construct a better class prototype as in Semi-ProtoNet (we denote this approach as {\feat}$^\ddagger$).

By using more unlabeled test instances in the transductive environment, {\feat}$^\dagger$ achieves further performance improvement compared with the standard {\feat}, which verifies the unlabeled instances could assist the embedding adaptation of the labeled ones. With more accurate class center estimation, {\feat}$^\ddagger$ gets a further improvement. The performance gain induced by the transductive \feat is more significant in the one-shot learning setting compared with the five-shot scenario, since the helpfulness of unlabeled instance decreases when there are more labeled instances.

\begin{table}[tbp]
	\small
	\tabcolsep 5pt
	\centering
	\caption{Results of generalized {\feat} with ConvNet backbone on {\it Mini}ImageNet. All methods are evaluated on instances composed by \textsc{seen} classes, \textsc{unseen}  classes, and both of them (\textsc{combined}), respectively.}
	\begin{tabular}{@{\;}lccc@{\;}}
		\addlinespace
		\toprule
		Measures $\rightarrow$ & \bf \textsc{seen} & \bf \textsc{unseen} & \bf \textsc{combined}\\ \midrule
		\multicolumn{4}{@{\;}l}{\bf 1-shot learning}\\
		ProtoNet                  & 41.73{\tiny $\pm$0.03} & 48.64{\tiny $\pm$0.20} &  35.69{\tiny $ \pm$0.03} \\
		{{\feat}}       & \bf 43.94{\tiny $\pm$0.03} & \bf 49.72{\tiny $\pm$0.20} & \bf 40.50{\tiny $\pm$0.03} \\
		\midrule
		\multicolumn{4}{@{\;}l}{\bf 5-shot learning}\\
		ProtoNet                  & 41.06{\tiny $\pm$0.03} & 64.94{\tiny $\pm$0.17} & 38.04{\tiny $ \pm$0.02} \\
		{{\feat}}       & \bf 44.94{\tiny $\pm$0.03} & \bf 65.33{\tiny $\pm$0.16} & \bf 41.68{\tiny $\pm$0.03} \\
		\midrule
		Random Chance & 1.56 & 20.00 & 1.45 \\
		\bottomrule
	\end{tabular}
	\label{supp-tab:joint}
\end{table}

\subsection{More Generalized FSL Results}
Here we show the full results of {\feat} in the generalized few-shot learning setting in Table~\ref{supp-tab:joint}, which includes both the 1-shot and 5-shot performance. All methods are evaluated on instances composed by \textsc{seen} classes, \textsc{unseen}  classes, and both of them (\textsc{combined}), respectively. In the 5-shot scenario, the performance improvement mainly comes from the improvement of over the \textsc{Unseen} tasks.

\subsection{Large-Scale Low-Shot Learning}
Similar to the generalized few-shot learning, the large-scale low-shot learning~\cite{HariharanG17Low,Gidaris2018Dynamic,Wang2018Low} considers the few-shot classification ability on both \textsc{seen} and \textsc{unseen} classes on the full ImageNet~\cite{RussakovskyDSKS15ImageNet} dataset. There are in total 389 \textsc{seen} classes and 611 \textsc{unseen} classes~\cite{HariharanG17Low}. We follow the setting (including the splits) of the prior work~\cite{HariharanG17Low} and use features extracted based on the pre-trained ResNet-50~\cite{he2016deep}.
Three evaluation protocols are evaluated, namely the top-5 few-shot accuracy on the \textsc{unseen} classes, on the combined set of both \textsc{seen} and \textsc{unseen} classes, and the calibrated accuracy on weighted by selected set prior on the combined set of both \textsc{seen} and \textsc{unseen} classes. The results are listed in Table~\ref{supp-tab:low_shot}. We observe that {\feat} achieves better results than others, which further validates {\feat}'s superiority in generalized classification setup, a large scale learning setup.

\begin{table}[tbp]
	\small
	\tabcolsep 5pt
	\centering
	\caption{The top-5 low-shot learning accuracy over all classes on the large scale ImageNet~\cite{RussakovskyDSKS15ImageNet} dataset (w/ ResNet-50).}
	\resizebox{\linewidth}{!}{
		\begin{tabular}{@{\;}cccccc@{\;}}
			\addlinespace
			\toprule
			{\sc Unseen} & 1-Shot   & 2-Shot     & 5-Shot     & 10-Shot    & 20-Shot \\
			\midrule
			{ProtoNet}~\cite{SnellSZ17Prototypical}    & 49.6  & 64.0  & 74.4  & 78.1  & 80.0 \\
			PMN~\cite{Wang2018Low}   & 53.3  & 65.2  & 75.9  & 80.1  & 82.6 \\
			\midrule
			{\feat}  & \bf 53.8  & \bf 65.4  & \bf 76.0  & \bf 81.2  & \bf 83.6 \\
			\midrule\midrule
			{All} & 1-Shot   & 2-Shot     & 5-Shot     & 10-Shot    & 20-Shot \\
			\midrule
			{ProtoNet}~\cite{SnellSZ17Prototypical}    & 61.4  & 71.4  & 78.0  & 80.0  & 81.1 \\
			PMN~\cite{Wang2018Low}   & 64.8  & 72.1  & 78.8  & 81.7  & 83.3 \\
			\midrule
			{\feat}  & \bf 65.1  & \bf 72.5  & \bf 79.3  & \bf 82.1  & \bf 83.9 \\
			\midrule\midrule
			{All w/ Prior} & 1-Shot   & 2-Shot     & 5-Shot     & 10-Shot    & 20-Shot \\
			\midrule
			{ProtoNet}~\cite{SnellSZ17Prototypical}    & 62.9  & 70.5  & 77.1  & 79.5  & 80.8 \\
			PMN~\cite{Wang2018Low}   & 63.4  & 70.8  & 77.9  & 80.9  & 82.7 \\
			\midrule
			{\feat}  & \bf 63.8  & \bf 71.2  & \bf 78.1  & \bf 81.3  & \bf 83.4 \\
			\bottomrule
		\end{tabular}
	}
	\label{supp-tab:low_shot}
\end{table}

\end{document}